%% file: main.tex
\documentclass[final,3p,times,twocolumn]{elsarticle}

\usepackage{amsmath}
\usepackage{amsfonts}
\usepackage{amssymb}
\usepackage{amsthm}
\usepackage{color}
\usepackage{multirow}
\usepackage{algorithm}
\usepackage{algpseudocode}
\usepackage{xspace}
\usepackage{makecell}
\usepackage{pifont}

\newcommand{\cmark}{\ding{51}\xspace}
\newcommand{\xmark}{\ding{55}\xspace}

\usepackage[most]{tcolorbox}

\newtcolorbox{takehome}[1]
{
	colback=red!3!white,
	colframe=red!75!white,  
	title={#1},
}
\usepackage{graphicx}
\usepackage[colorlinks=true, allcolors=blue]{hyperref}
\input{paper/header}

\journal{Annual Reviews in Control}

\begin{document}

\begin{frontmatter}



\title{A survey on secure decentralized optimization and learning}


\author[KTH]{Changxin Liu}
\ead{changxin@kth.se}
\author[KTH]{Nicola Bastianello}
\ead{nicolba@kth.se}
\author[HKUST]{Wei Huo}
\ead{whuoaa@connect.ust.hk}
\author[UVic]{Yang Shi}
\ead{yshi@uvic.ca}
\author[KTH]{Karl H. Johansson}
\ead{kallej@kth.se}

\affiliation[KTH]{organization={Division of Decision and Control Systems, KTH Royal Institute of Technology, and Digital Futures}, city={Stockholm}, country={Sweden}}
\affiliation[HKUST]{organization={Department of Electronic and Computer Engineering, Hong Kong University of Science and Technology}, addressline={Clear Water Bay, Kowloon}, city={Hong Kong SAR},country={China}}
\affiliation[UVic]{organization={Department of Mechanical Engineering, University of Victoria}, city={Victoria}, state={BC}, country={Canada}}
%
%

\begin{abstract}
Decentralized optimization has become a standard paradigm for solving large-scale decision-making problems and training large machine learning models without centralizing data. However, this paradigm introduces new privacy and security risks, with malicious agents potentially able to infer private data or impair the model accuracy. Over the past decade, significant advancements have been made in developing secure decentralized optimization and learning frameworks and algorithms. This survey provides a comprehensive tutorial on these advancements. We begin with the fundamentals of decentralized optimization and learning, highlighting centralized aggregation and distributed consensus as key modules exposed to security risks in federated and distributed optimization, respectively. Next, we focus on privacy-preserving algorithms, detailing three cryptographic tools and their integration into decentralized optimization and learning systems. Additionally, we examine resilient algorithms, exploring the design and analysis of resilient aggregation and consensus protocols that support these systems. We conclude the survey by discussing current trends and potential future directions.

\end{abstract}



\end{frontmatter}

 \tableofcontents

\input{paper/introduction}
\input{paper/preliminaries}
\input{paper/private_DO}
\input{paper/robust_DO}

\input{paper/conclusion}

\section*{Acknowledgements}
This work was partially supported by the European Union’s Horizon Research and Innovation Actions programme under grant agreement No. 101070162, partially by the Swedish Research Council Distinguished Professor Grant 2017-01078, the Knut and Alice Wallenberg Foundation Wallenberg Scholar Grant, and partially by the NSERC Postdoctoral Fellowship PDF-577876-2023.

\bibliographystyle{elsarticle-num}
\bibliography{survey}

\end{document}

%% file: paper/header.tex
\usepackage{color}
\usepackage{amsmath}
\usepackage{amssymb}
\usepackage{amsthm}
\usepackage{caption}
\usepackage{subcaption}
\usepackage{mathtools}
\usepackage{appendix}
\usepackage{tabularx,colortbl}
\usepackage{tikz}
\usepackage[nice]{nicefrac}

\newtheorem{lemma}{Lemma}

\newtheorem{assumption}{Assumption}

\theoremstyle{definition}
\newtheorem{definition}{Definition}
\newtheorem{example}{Example}

\newcommand{\N}{\mathbb{N}}
\newcommand{\R}{\mathbb{R}}

\newcommand{\CommentState}[1]{\Statex\hspace{\algorithmicindent}{\color{blue}// #1}}

\newcommand{\sz}[1]{\left|#1\right|}

\renewcommand{\emptyset}{\varnothing}

\def\expec#1#2{{\mathbb{E}}_{#1}\left[ #2 \right]}

\newcommand{\norm}[1]{\left\lVert#1\right\rVert}

\newcommand{\calH}{\mathcal{H}}

\newcommand{\calD}{\mathcal{D}}
\newcommand{\calG}{\mathcal{G}}
\newcommand{\calL}{\mathcal{L}}

\newcommand{\calV}{\mathcal{V}}

\newcommand{\calC}{\mathcal{C}}

\newcommand{\calN}{\mathcal{N}}

\newcommand\xx{\boldsymbol{\mathit{x}}}
\newcommand\ff{\boldsymbol{\mathit{f}}}
\newcommand\yy{\boldsymbol{\mathit{y}}}

\newcommand\ww{\boldsymbol{\mathit{w}}}

\renewcommand\AA{\boldsymbol{\mathit{A}}}

\renewcommand{\emptyset}{\varnothing}

\DeclareMathOperator*{\argmin}{argmin}  

\definecolor{lightGreen}{RGB}{150,255,150}

%% file: paper/introduction.tex
\section{Introduction}

\label{sec:intro}

The confluence of several technological trends in recent years has revolutionized the training of high-performance machine learning models.
Indeed, connected devices are increasingly widespread on the customer and industrial markets, allowing to collect massive amounts of data. These devices are also equipped with growing computational power, enabling efficient model training.
However, data privacy concerns and regulatory restrictions, such as the General Data Protection Regulation (GDPR) \cite{GDPR2016a} and the EU AI Act \cite{eu2023aiact} in Europe,
make the traditional approach of consolidating data into a single repository undesirable, and often also impractical.

The goal of training efficient models without compromising privacy has thus spurred the adoption of decentralized algorithms.
In this paradigm, spatially distributed computing units, utilizing wireless communication networks and cloud-based computing platforms, collaboratively train machine learning models without disclosing local private data. Decentralized learning algorithms are employed in an increasingly wide range of applications. Examples include the training of predictive keyboards for smartphones by Google \cite{hard2018federated}; training diagnostic models for healthcare leveraging multi-institution collaborations \cite{rauniyar_federated_2023}; intelligent transportation \cite{zhang_federated_2024,qu_decentralized_2021}; and many more, see \cite{li_review_2020} for a survey.

Besides regulatory constraints to safeguard privacy, we remark that adopting the decentralized paradigm also yields a practical advantage.
Indeed, tasks like training foundational models with billions of parameters could take years of computation if performed on a single machine \cite{lidemystifying}. 
To significantly reduce training time, distributed algorithms leveraging multiple parallel computing nodes are a natural choice \cite{goyal2017accurate}. For instance, DeepMind developed a massively distributed architecture for training deep Q-networks (DQNs) for reinforcement learning \cite{nair2015massively}, which achieves comparable performance but ten times faster than single-GPU DQN implementations for most Atari games \cite{mnih2015human}.
Decentralized optimization algorithms have also been employed in data centers to accelerate training \cite{mikami2018massively}.



In recent years, research on decentralized optimization and learning algorithms has thus flourished, and we reference the surveys \cite{li_federated_2020,gafni_federated_2022,nedic2018distributed,yang_survey_2019} for a comprehensive overview.
However, alongside these advancements, the risk of cyber-attacks on decentralized systems has significantly increased. This is especially relevant in sensitive applications such as healthcare and transportation.
Referencing the computer security literature, there are three fundamental security properties of information technology systems: \textit{confidentiality}, \textit{integrity}, and \textit{availability},  often referred to as the CIA triad \cite{matt2002computer,teixeira2015secure}.
However, \textit{employing the decentralized paradigm is not enough to satisfy the CIA properties}, showing, \textit{e.g.}, a vulnerability to data privacy attacks despite not disclosing data \cite{zhu2020deep}.
The situation is further complicated by ubiquitous cyber-channels in large-scale systems, which are inherently susceptible to adversarial attacks compromising parts of the system \cite{su2016fault}. These attacks can leak sensitive information from datasets and impair the performance of decision-making or machine learning models. As more centralized learning systems transition to decentralized models, the importance of secure decentralized optimization and learning algorithms becomes increasingly critical.

\begin{figure}[t]
\centering
    \includegraphics[scale=0.65]{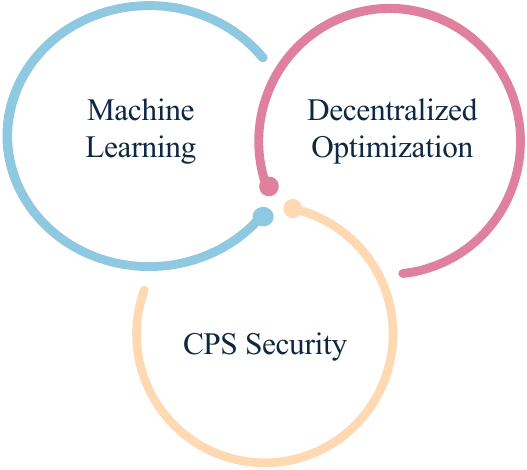}
\caption{The intersection of machine learning, decentralized optimization, and CPS security.}
\label{fig:intersection}
\end{figure}

Addressing this pressing issue requires efforts across three domains: machine learning, decentralized optimization, and cyber-physical system (CPS) security, as illustrated in Figure \ref{fig:intersection}. Recently, significant advancements have been made in developing secure decentralized optimization and learning algorithms. In this paper, we provide a comprehensive review of recent developments in the field.

\subsection{Organization of the paper}

The survey paper is organized around the two main classes of attacks that secure decentralized optimization and learning aim to address:

\begin{itemize}
    \item \textit{Attacks on data privacy}: In these attacks, the attackers attempt to steal sensitive information from other honest participants. The attackers, whether inside or outside the system, may record intermediate model updates and gradients. These attacks can be further classified as feature or attribute inference attacks \cite{hitaj2017deep,melis2019exploiting}, membership inference attacks \cite{shokri2017membership,nasr2019comprehensive}, data reconstruction attacks \cite{zhu2020deep}. Particularly for data reconstruction attacks, advanced inference techniques based on the minimization of gradient difference measured in distance \cite{zhu2020deep} and cosine similarity \cite{geiping2020inverting} were reported.

    \item \textit{Attacks on decision or model security}: In these attacks, the attackers aim to impair the performance of the final decision or model by maliciously manipulating local data and updates. Examples include data poisoning attacks \cite{bhagoji2019analyzing} and model poisoning attacks \cite{baruch2019little}. While arbitrarily behaved poisoning attacks are easily detectable, strategies have been developed to make the poisoning stealthier \cite{bagdasaryan2020backdoor}. Note that the free-riding attack \cite{lin2019free}, where attackers get the global decision or model without contributing to the process (\textit{e.g.}, by uploading random updates), can also be considered a form of the model poisoning attack.
\end{itemize}


\begin{figure*}[t]
\centering
    \includegraphics[scale=0.6]{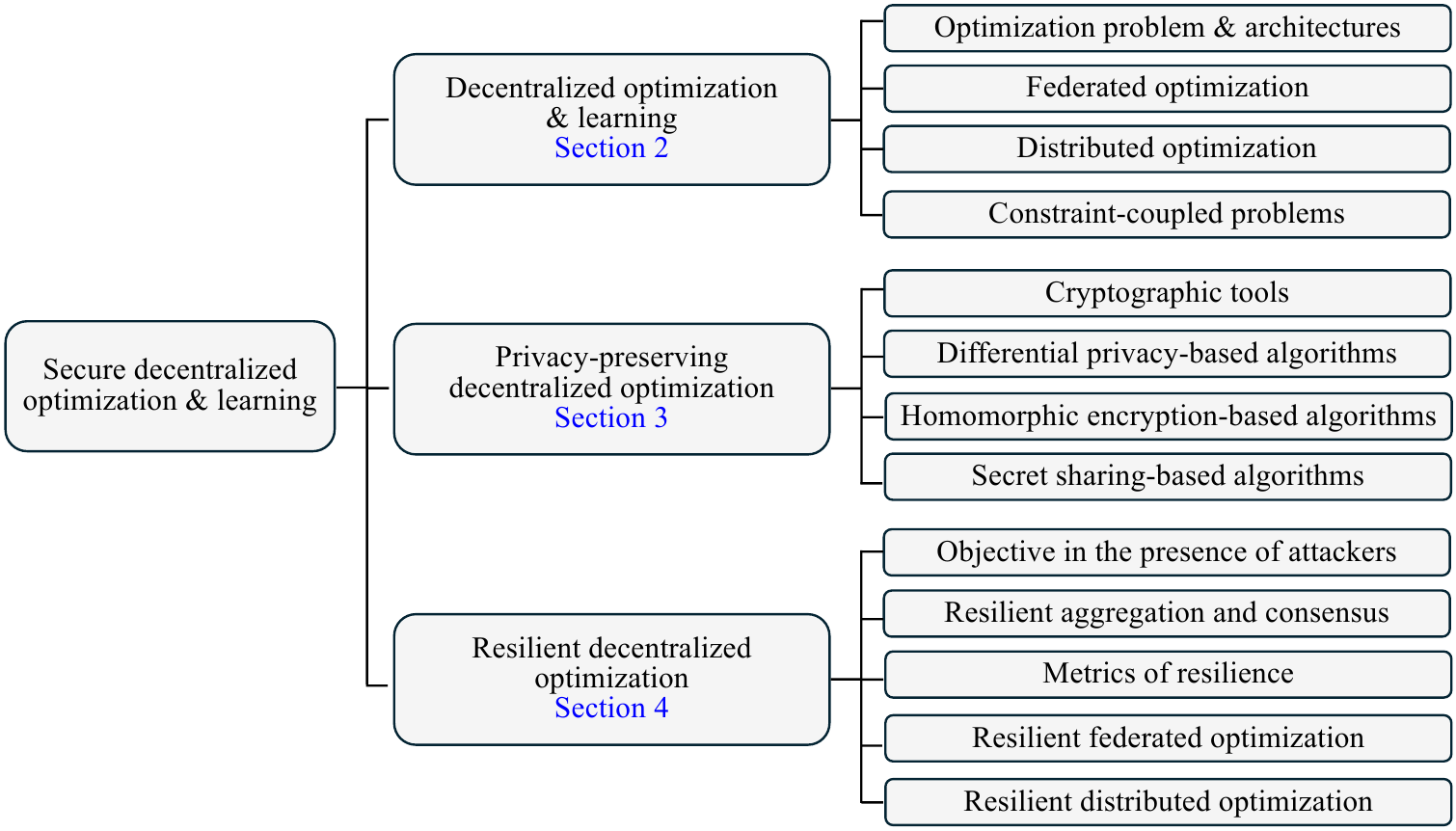}
\caption{Organization of the paper.}
\label{fig:ToC}
\end{figure*}


The outline of the paper, depicted in Figure \ref{fig:ToC}, is as follows.
We begin by presenting preliminaries on decentralized optimization and learning over networks in Section \ref{sec:preliminaries}. Section \ref{sec:private-optimization} focuses on privacy-preserving decentralized algorithms designed to protect against data privacy attacks, providing a tutorial on three cryptographic tools and their integration into decentralized optimization and learning algorithms. Then, in Section \ref{sec:resilient-optimization}, we survey resilient decentralized optimization and learning algorithms. To offer an in-depth understanding of the underlying principles, we elaborate on the design and analysis of resilient aggregation and consensus protocols, which are key enablers of resilient decentralized algorithms. Section \ref{sec:conclusion} concludes the survey with discussions on potential future directions.

\subsection{Related survey papers}

There are other surveys dedicated to the secure implementation of decentralized optimization and learning in CPS. Different from this survey, they primarily focus on a limited range of attacks, such as privacy-preserving decentralized optimization \cite{chen2024privacy} and resilient decentralized optimization \cite{yang2020adversary,guerraoui2023byzantine}. We comprehensively address adversarial attacks on both data privacy and model security. Moreover, the aim of this survey is not only to document the state-of-the-art secure solutions to these problems, but also to provide a tutorial on their development. This approach will facilitate an in-depth understanding of the underlying principles of secure decentralized optimization and learning. We also highlight upcoming trends in the field.

%% file: paper/preliminaries.tex
\paragraph{Notation}
\def\prox{\operatorname{prox}}
\def\proj{\operatorname{proj}}
$\mathbb{R}$, $\mathbb{ R}^d$, and $\mathbb{R}^{d\times d}$, for some given $d>0$, denote the set of real numbers, vectors, and matrices, respectively. 
$\N$ and $\mathbb{Z}$ denote the set of natural numbers and integers, respectively. 
Given $n \in \N$, we denote by $[n]$ the set of integers $\{1, 2, \dots, n \}$.
Given $q\in\N$ and $\nu \in\mathbb{Z}$, the modulo operation is defined as $\nu \,\, \text{mod} \,\, q: = \nu - \lfloor{\nicefrac{\nu}{q}} \rfloor q$, where $\lfloor \cdot \rfloor$ denotes the floor function. Column vectors are considered as the default orientation unless otherwise stated. 
The symbol $\geq$ is element-wise when applied to vectors.
All norms $\lVert \cdot \rVert$ are $2$-norms unless otherwise stated. For a vector $x\in \mathbb{R}^d$ and a closed convex set $\mathcal{X}\subseteq \mathbb{R}^d$, the orthogonal projection mapping is defined as $\proj_{\mathcal{X}}[x]:= \argmin_{y\in\mathcal{X}} \lVert y-x \rVert$. Let $\otimes$ be the Kronecker product. The cardinality of a set $S$ is denoted as $\lvert S \rvert$.

The notation $\mathcal{O}(\cdot)$ is used to represent the asymptotic upper bound on the growth rate of a function. Specifically, $h(n) = \mathcal{O}(g(n))$ means that there exist positive constants $C$ and $n_0$ such that $0 \leq h(n) \leq Cg(n)$ for all $n \geq n_0$. 
The notation $\Theta(\cdot)$ represents the tightest asymptotic bound. $h(n)$ is said to be $\Theta(g(n))$ if $h(n)$ is bounded above and below by constant multiples of $g(n)$ for sufficiently large values of $n$.


To describe the network topology, we denote by $\mathcal{G}=(\mathcal{V},\mathcal{E})$ a \textit{directed graph} (in short, \textit{digraph}), where $\mathcal{V}=[n]$ denotes the set of $n$ nodes (or agents) and $\mathcal{E}\subseteq \mathcal{V}\times\mathcal {V}$ represents the set of edges. 
For $i,j\in\mathcal{V}$, 
the ordered pair $(i,j)\in \mathcal{E}$ denotes an edge from $i$ to $j$.
Node $j$ is said to be an in-neighbor of $i$ if $(j,i)\in\mathcal{E}$, 
and the set of $i$'s in-neighbors is denoted by $ \mathcal{N}_{i}^-=\{j\in \mathcal{V}|(j,i)\in \mathcal{E} \}$. Similarly, the set of $i$'s out-neighbors is defined as $ \mathcal{N}_{i}^+=\{j\in \mathcal{V}|(i,j)\in \mathcal{E}\}$. A directed path in a graph is an ordered sequence of nodes such that any ordered pair of nodes appearing consecutively in the sequence is an edge. If there exists a directed path between $i$ and $j$, then $j$ is said to be reachable from $i$. A graph is \textit{strongly connected} if every node is reachable from any other node. A graph $\mathcal{G}$ is \textit{undirected} if $(i,j)\in \mathcal{E}$ implies that $(j,i)\in\mathcal{E}$.

\section{Decentralized optimization and learning}\label{sec:preliminaries}
The traditional centralized optimization and learning paradigm, depicted in Figure~\ref{fig:architecture-centralized}, entails collecting data in a single location, where it is then processed. The effectiveness and performance of this set-up has been demonstrated extensively.
However, in many applications these data are sensitive in nature, and transmitting them would potentially expose them to malicious agents. Moreover, recent technological advances have enabled the aggregation of huge volumes of data, whose transmission would be resource intensive and impractical.
A paradigm shift is therefore needed to overcome the limitations of the centralized approach. The objective then becomes that of \textit{enabling cooperative optimization and learning without the agents sharing raw data}.

\subsection{Optimization problem and system architectures}\label{subsec:architectures}



\begin{figure*}[!ht]
\centering
\begin{subfigure}{.333\textwidth}
\centering
    \includegraphics[scale=0.65]{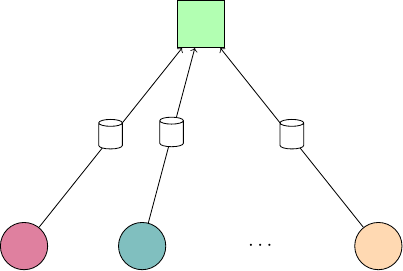}
    \caption{Centralized}
    \label{fig:architecture-centralized}
\end{subfigure}%
\begin{subfigure}{.333\textwidth}
\centering
    \includegraphics[scale=0.65]{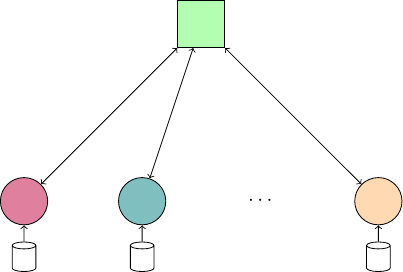}
    \caption{Federated}
    \label{fig:architecture-federated}
\end{subfigure}%
\begin{subfigure}{.333\textwidth}
\centering
    \includegraphics[scale=0.65]{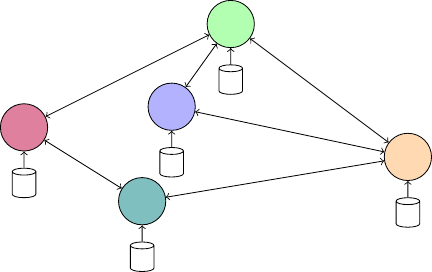}
    \caption{Distributed}
    \label{fig:architecture-distributed}
\end{subfigure}
\caption{Depictions of the three main architectures for optimization and learning.}
\label{fig:architectures}
\end{figure*}

This section provides a concise introduction to decentralized optimization and learning. The objective is to highlight fundamental concepts and challenges, while for an in-depth literature review we reference \cite{nedic2018distributed,yang_survey_2019,notarstefano_distributed_2019,li_federated_2020,zhang_survey_2021,gafni_federated_2022,gao_when_2023}. We use several interchangeable terms commonly found in the learning and optimization community, such as objective, cost, or loss, and solution, model, or decision.

We start formulating the problem at hand through empirical risk minimization, a foundational technique in learning.


\begin{example}[Empirical risk minimization]
\label{em:ERM}
Consider a set of $n \in \N$ agents, each locally storing a private dataset $\{ \xi_i^h \}_{h = 1}^{m_i}$, $i \in \{ 1, \ldots, n \}$, $m_i \in \N$, $\xi_i^h \in \R^q$. Given a loss function $\ell : \R^d \times \R^q \to \R : (x, \xi) \mapsto \ell(x, \xi)$, which identifies the accuracy of a model, we can define the \textit{empirical risk minimization} (ERM) problem \cite{mohri_foundations_2018}
\begin{equation}\label{eq:centralized-erm}
    \min_{x \in \mathbb{R}^d} \,\, \left\{ \frac{1}{n}\sum_{i=1}^n \sum_{h = 1}^{m_i} \ell(x, \xi_i^h) \right\}.
\end{equation}
As an example, we can take $\xi_i^h = (a_i^h, b_i^h) \in \R^{1 \times d} \times \{ -1, 1 \}$ and $\ell(x, \xi_i^h) = \log\left( 1 + \exp\left( - b_i^h a_i^h x \right) \right)$ for a classification task. In this task, we aim to minimize the fitting error of a logistic regression model parameterized by $x$. The goal is to optimize the model's parameters so that it can effectively classify input data into discrete categories.
If the local datasets are collected in a single location according to the centralized model of Figure~\ref{fig:architecture-centralized}, the problem can be solved as a traditional ERM \cite{mohri_foundations_2018}. However, as discussed above, we want to resort to decentralized architectures to avoid sharing raw data.
\end{example}

Abstracting now from the example of ERM, our objective is to enable the cooperative solution of the class of problems
\begin{equation}\label{eq:centralized-optimization}
    \min_{x \in \mathbb{R}^d} \,\, \left\{ f(x)=\frac{1}{n}\sum_{i=1}^n f_i(x)  \right\},
\end{equation}
without sharing the local cost (or loss) functions $f_i : \R^d \to \R$, which may otherwise disclose private data.
The idea is to reformulate~\eqref{eq:centralized-optimization} into the following \textit{consensus optimization} problem \cite{nedic2018network,gafni_federated_2022}
\begin{equation}\label{eq:decentralized-optimization}
    \min_{x_i \in \mathbb{R}^d, \ \forall i} \,\, \left\{\frac{1}{n}\sum_{i=1}^n f_i(x_i)  \right\} \qquad
   \text{s.t.} \quad x_1 = \ldots = x_n,
\end{equation}
whose solutions are denoted
$$
    \xx^* = [ x^* ]_{i = 1}^n, \qquad x^* \in \argmin_{x \in \R^d} \frac{1}{n} \sum_{i = 1}^n f_i(x).
$$
Let $\xx = [ x_i ]_{i = 1}^n$ and define the \textit{consensus constraint set} $\calC = \{ \xx \in \mathbb{R}^{nd} \ | \ x_1 = \ldots = x_n \}$.
We make the following standing assumption to simplify our discussion.

\begin{assumption}[Smoothness]\label{as:smoothness}
The local cost functions are $L$-smooth:
$$
    \norm{\nabla f_i(x) - \nabla f_i(y)} \leq L \norm{x - y}, \quad \forall x, y \in \mathbb{R}^d, \ i \in [n].
$$
\end{assumption}

\smallskip

Let us now discuss the solution of~\eqref{eq:decentralized-optimization} under Assumption~\ref{as:smoothness}.
By smoothness, we can apply the \textit{projected gradient descent} \cite{bauschke_convex_2017}, which yields $\xx^{(t+1)} = \proj_{\calC}\left( \xx^{(t)} - \alpha \nabla \ff(\xx^{(t)}) \right)$, where $\nabla \ff(\xx)$ is a vector stacking the local gradients $\nabla f_i(x_i)$.
Since the projection of a vector $\xx$ onto the consensus set corresponds to averaging its components, the projected gradient descent can be rewritten as the following updates, $t \in \N$, $i \in [n]$:
\begin{subequations}\label{eq:distributed-pgd}
\begin{align}
    y_i^{(t+1)} &= x_i^{(t)} - \alpha \nabla f_i(x_i^{(t)}) \label{eq:distributed-pgd-1} \\
    x_i^{(t+1)} &= \frac{1}{n} \sum_{j = 1}^n y_j^{(t+1)}. 
    \label{eq:distributed-pgd-2-bis}
\end{align}
\end{subequations}
Update~\eqref{eq:distributed-pgd-1} is a \textit{local training} step, during which each agent performs a gradient step on the local cost. Agents only need access to local information (\textit{e.g.}, their dataset $\{ \xi_i^h \}_{h = 1}^{m_i}$) to perform this update.
Update~\eqref{eq:distributed-pgd-2-bis} instead performs \textit{aggregation} of the results of local training, and requires cooperation from all agents. Depending on the architecture deployed to solve~\eqref{eq:decentralized-optimization}, there are different ways of implementing this aggregation, which we will discuss in  Sections~\ref{subsubsec:federated-opt}~and~\ref{subsubsec:distributed-opt}.



We distinguish two main architectures to enable decentralized optimization and learning, namely \textit{federated}, depicted in Figure~\ref{fig:architecture-federated}, and \textit{distributed}, depicted in Figure~\ref{fig:architecture-distributed}.
The main difference lies in how the agents that store data are interconnected. In federated architectures, the agents are aided by a \textit{coordinator}, which communicates with all of them and aggregates their local results into a more accurate solution or model.
In distributed learning, instead, agents only rely on peer-to-peer communications to converge to a global solution or model.

More specifically, we use the following terminology:
\begin{itemize}
    \item \textit{Centralized}: to reference traditional set-ups in which data is collected at a single location for processing;
    
    \item \textit{Decentralized}: to reference any set-up in which data is privately stored by a set of agents, which process it locally and collaboratively share their results.
\end{itemize}
Under the decentralized umbrella, we further differentiate:
\begin{itemize}
    \item \textit{Federated}: in which the coordinator serves the special role of aggregating the results of local computations by communicating with all agents;
    
    \item \textit{Distributed} (also called \textit{peer-to-peer} or \textit{decentralized federated}): in which the agents are connected to each other according to some communication graph topology, and no agent serves a privileged role.
\end{itemize}

\subsection{Federated optimization}\label{subsubsec:federated-opt}
The federated architecture of Figure~\ref{fig:architecture-federated} is composed of a coordinator connected to each of the $n$ agents. Exploiting this structure, it is therefore possible to implement~\eqref{eq:distributed-pgd} as detailed in Algorithm~\ref{alg:federated-gradient-naive}.
\begin{algorithm}[t]
\caption{Na\"ive federated gradient descent}
\label{alg:federated-gradient-naive}
\begin{algorithmic}[1]
	\Require initial conditions $x_i^{(0)}$, $i \in [n]$, step-size $\alpha > 0$.
 
	\For{$t = 0,1,\ldots$ every agent $i$}
	\CommentState{local training}
	\State each agent $i$ performs~\eqref{eq:distributed-pgd-1}
    $
        y_i^{(t+1)} = x_i^{(t)} - \alpha \nabla f_i(x_i^{(t)})
    $
    \State and transmits the result to the coordinator
	\CommentState{aggregation}
	\State the coordinator collects $\{ y_i^{(t+1)} \}_{i = 1}^n$, aggregates them
    $
        y^{(t+1)} = \frac{1}{n} \sum_{i = 1}^n y_i^{(t+1)}
    $
    and broadcasts $y^{(t+1)}$ to all agents, which set $x_i^{(t+1)} = y^{(t+1)}$
	\EndFor
\end{algorithmic}
\end{algorithm}
In principle, Algorithm~\ref{alg:federated-gradient-naive} is sufficient to cooperatively solve~\eqref{eq:decentralized-optimization}.
However, deploying this algorithm in practice faces several challenges \cite{li_federated_2020,gafni_federated_2022}, which require us to refine its design.
In the following, we discuss these challenges one by one, and the design solutions that have been proposed to address them.
We defer the treatment of the challenge of security to Sections~\ref{sec:private-optimization} and~\ref{sec:resilient-optimization}.

\paragraph{Communications bottleneck}
Each iteration of Algorithm~\ref{alg:federated-gradient-naive} includes a round of communications from all agents to the coordinator (to transmit $\{ y_i^{(t+1)} \}_{i = 1}^n$), and then back from the coordinator to the agents ($y^{(t+1)}$).
However, the network carrying these communications may act as a bottleneck. On the one hand, the available bandwidth may be limited when employing wireless (or over-the-air) communications \cite{qian_distributed_2022}.
On the other, the size of the packets being shared may be large ($d \gg 1$) especially when training high-dimensional models such as (deep) neural networks.
The goal therefore is to \textit{reduce the communication requirements} of Algorithm~\ref{alg:federated-gradient-naive}.

Different techniques have been leveraged to this end, which we summarize in the following:
\begin{itemize}
    \item \textit{Reducing the size} of communications: by applying \textit{quantization or compression}. A host of different quantization/communication techniques have been proposed \cite{xu2024quantized}, and we reference \cite{zhao_towards_2023} for a comprehensive overview.

    \item \textit{Reducing the number} of communications: via partial participation, local training, or event-triggered communication. \textit{Partial participation}, or \textit{client selection}, reduces the number of agents that communicate with the coordinator at each iteration. 
    The complementary approach of \textit{local training}, instead, consists in performing multiple steps (or epochs) of local training per each communication round \cite{grudzien_can_2023}.
Another approach is \textit{event-triggered} communication, which initiates communication only when necessary \cite{xu2023distributed,liu2019distributed}.
\end{itemize}

\paragraph{Resources heterogeneity}
In many applications, the agents cooperating towards the solution of~\eqref{eq:decentralized-optimization} are equipped with highly heterogeneous resources \cite{li_review_2020}. For example, they may rely on different computational powers, communication and storage capabilities, as well as different battery powers.
We can thus identify two consequences of resource heterogeneity:
\begin{itemize}
    \item \textit{Asynchronous local computations}: agents with different computational powers, or different power consumption rates, may conclude a round of local training with different speeds, thus communicating more or less frequently with the coordinator \cite{li_review_2020}. This results in partial participation, albeit due to external constraints rather than by design choice.

    \item \textit{Inexact local training}: to account for their limited resources, the agents may resort to inexact local training techniques. The foremost example is the computation of \textit{stochastic gradients} $\hat{\nabla} f_i$ rather than full gradients $\nabla f_i$ \cite{gorbunov_local_2021}.
    This may however result in loss of accuracy.
\end{itemize}

\smallskip

We are now ready to re-design Algorithm~\ref{alg:federated-gradient-naive} by incorporating these techniques. The resulting Algorithm~\ref{alg:fed-avg} coincides with \textit{FedAvg}, the foundational federated algorithm proposed in \cite{mcmahan_communication_2017}.
\begin{algorithm}[t]
\caption{FedAvg}
\label{alg:fed-avg}
\begin{algorithmic}[1]
	\Require initial conditions $x_i^{(0)}$, $i \in [n]$, step-size $\alpha > 0$, number of participating clients $n_\mathrm{p}$, local training epochs $n_\mathrm{e}$.
 
	\For{$t = 0,1,\ldots$}
	\CommentState{local training}
    \State randomly pick $n_\mathrm{p}$ participating agents 
	\For{each participating agent $i$}
    \State set $w_{i,0}^{(t)} = x_i^{(t)}$ and
    \For{$k = 0, 1, \ldots, n_\mathrm{e}$} 
    \State 
    $
        w_{i,k+1}^{(t)} = w_{i,k}^{(t)} - \alpha \hat{\nabla} f_i(w_{i,k}^{(t)})
    $ 
    \EndFor
    \State set $y_i^{(t+1)} = w_{i,n_\mathrm{e}}^{(t)}$ and transmit it to the coordinator 
    \EndFor
    
	\CommentState{aggregation}
	\State the coordinator collects $\{ y_i^{(t+1)} \}$ from the participating agents, aggregates them
    $$
        y^{(t+1)} = \frac{1}{n} \left( \sum_{\text{$i$ part.}} y_i^{(t+1)} + \sum_{\text{$i$ not part.}} y_i^{(t)}\right)
    $$
    and broadcasts $y^{(t+1)}$ to all agents, which set $x_i^{(t+1)} = y^{(t+1)}$ 
	\EndFor
\end{algorithmic}
\end{algorithm}

\paragraph{One last challenge: statistical heterogeneity}
Recall the empirical risk minimization problem~\eqref{eq:centralized-erm}, where each agent collects and stores a dataset $\{ \xi_i^h \}_{h = 1}^{m_i}$, which defines the local cost
$
    f_i(x) = \sum_{h = 1}^{m_i} \ell(x, \xi_i^h).
$
The datapoints of each agent can be modeled as randomly drawn from a given distribution $\xi_i^h \sim \calD_i$. However, in most applications these \textit{data distributions are heterogeneous} \cite{gafni_federated_2022}.

\begin{figure}[t]
\centering
    \includegraphics[scale=1]{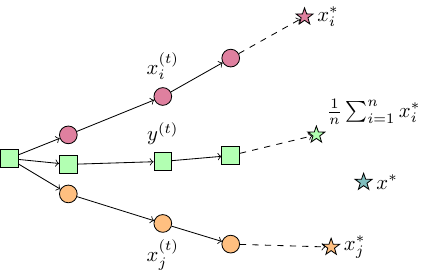}
\caption{An illustration of client drift.}
\label{fig:client-drift}
\end{figure}

As discussed above, the design of Algorithm~\ref{alg:fed-avg} includes the use of multiple local training steps to reduce the number of communications. However, this heuristic turns out to be detrimental when the data distributions $\{ \calD_i \}_{i = 1}^n$ are heterogeneous (or non-i.i.d.).
Indeed, during local training an agent's solution is skewed by the biased perspective that its data offers, leading to \textit{client drift} \cite{karimireddy_scaffold_2020}, depicted in Figure~\ref{fig:client-drift}.
In particular, the trajectory $\{ x_i^{(t)} \}_{t \in \N}$ of each agent $i$ converges towards a local solution $x_i^* \in \argmin_{x \in \R^d} f_i(x)$. And the trajectory $\{ y^{(t)} \}_{t \in \N}$ generated by the coordinator converges towards $\frac{1}{n} \sum_{i = 1}^n x_i^*$ rather than the true solution $x^*$.

Research into solutions to statistical heterogeneity has proliferated in the past years, with \cite{grudzien_can_2023} detailing a short history. The main goal is the design of novel federated algorithms which employ multiple local training steps without undergoing client drift. In other words, local training is no longer used as a heuristic, with provable guarantees on the algorithms' convergence.


    
    
    

    

\subsection{Distributed optimization}\label{subsubsec:distributed-opt}
The algorithms discussed in Section~\ref{subsubsec:federated-opt} were designed on the blueprint of the projected gradient descent in \eqref{eq:distributed-pgd},
where the coordinator implements the projection~\eqref{eq:distributed-pgd-2-bis} onto the consensus set.
However, distributed architectures in general lack a central node capable of averaging local computations. The question then is: \textit{how can a projection onto the consensus set be implemented in a distributed fashion?}

In the following we briefly discuss the research stemming from this question, and discuss how the challenges of Section~\ref{subsubsec:federated-opt} also impact the design of distributed algorithms. To simplify our notation, we consider undirected graphs only.

\paragraph{Distributed gradient descent}
Research on the design of distributed protocols that converge to a consensus on the average of local quantities $\{ u_i \}_{i = 1}^n$ has matured significantly in the past decades \cite{kia_tutorial_2019}.
The central protocol that has been developed and analyzed is that of \textit{average consensus}:
\begin{equation}\label{eq:average-consensus}
    z_{i,0} = u_i, \quad
    z_{i,k+1} = \sum_{j \in \calN_i \cup \{ i \}} w_{ij} z_{j,k} \qquad k = 0, 1, \ldots
\end{equation}
where $\calN_i$ is the set of agents that exchange information with $i$, and $w_{ij}$ are positive coefficients that sum to one.

Therefore, one of the first attempts at designing a distributed version of~\eqref{eq:distributed-pgd} has been to try and implement the projection~\eqref{eq:distributed-pgd-2-bis} with average consensus~\eqref{eq:average-consensus}. The resulting distributed gradient descent algorithm is \cite{chen_diffusion_2012}
\begin{subequations}\label{eq:dgd}
\begin{align}
    y_i^{(t+1)} &= x_i^{(t)} - \alpha \nabla f_i(x_i^{(t)}) \\
    x_i^{(t+1)} &= \sum_{j \in \calN_i \cup \{ i \}} w_{ij} y_j^{(t+1)}.
\end{align}
\end{subequations}
However, this algorithm does not converge to the optimal solution: in fact, it can only reach a neighborhood thereof, whose radius depends on the magnitude of the step-size $\alpha$ \cite{chen_diffusion_2012}, see also \cite{yuan_convergence_2016,sundararajan_canonical_2019}.
As such, exact -- but sub-linear -- convergence can only be achieved by employing a diminishing step-size \cite{nedic2018network}.

\paragraph{Gradient tracking}
The issue with~\eqref{eq:dgd} is that the consensus projection~\eqref{eq:distributed-pgd-2-bis} is applied on a dynamic quantity, which changes at every local gradient step.
It is therefore necessary to resort to distributed protocols that can track the dynamic average of time-varying signals.
This gave rise to the class of \textit{gradient tracking} algorithms.

Let $\{ u_{i,k} \}_{i = 1}^n$, $k \in \N$ be local signals, then one example of dynamic average consensus is \cite{kia_tutorial_2019}
\begin{equation}\label{eq:dynamic-average-consensus}
    z_{i,k+1} = \sum_{j \in \calN_i \cup \{ i \}} w_{ij} \left( z_{j,k} + u_{j,k+1} - u_{j,k} \right) \qquad k = 0, 1, \ldots.
\end{equation}
Applying this scheme to approximate the consensus projection~\eqref{eq:distributed-pgd-2-bis} thus yields the algorithm
\begin{subequations}\label{eq:gt}
\begin{align}
    y_i^{(t+1)} &= x_i^{(t)} - \alpha \nabla f_i(x_i^{(t)}) \\
    x_i^{(t+1)} &= \sum_{j \in \calN_i \cup \{ i \}} w_{ij} \left( x_j^{(t)} + y_j^{(t+1)} - y_j^{(t)} \right)
\end{align}
\end{subequations}
which converges to the exact solution with a constant step-size \cite{yuan_exact_2019}. For a comprehensive overview of gradient tracking algorithms we reference \cite{nedic2018network,xin_general_2020}.

\paragraph{Distributed dual averaging}
Let $d(\cdot):\R^d \rightarrow \mathbb{ R}$ be a strongly convex proximal function, then applying the (centralized) dual averaging to~\eqref{eq:centralized-optimization} yields the update \cite{nesterov2009primal}:
\begin{equation}\label{centralized_DA_2}		{x}^{(t+1)}=\argmin_{x\in\R^d}\left\{\alpha_t \left\langle \sum_{\tau=0}^{t}g^{(\tau)}, x \right\rangle + d(x) \right\}
\end{equation}
where $g^{(t)}$ is a subgradient of $\frac{1}{n} \sum_{i \in \calV} f_i$ at $x^{(t)}$, and $\alpha_t $ is a non-increasing sequence of positive parameters.
In particular, we can choose
\begin{equation*}
	d(x) = \tilde{d}(x)-\tilde{d}(x^{(0)}) - \langle \nabla \tilde{d}(x^{(0)}), x-x^{(0)} \rangle,
\end{equation*} 
where $\tilde{d}$ is any $1$-strongly convex function, such as $\tilde{d}(x)=\lVert x \rVert^2/2$.

Similarly to the consensus projection~\eqref{eq:distributed-pgd-2-bis}, we need to design a protocol that can compute $z^{(t)} = \sum_{\tau=0}^{t}g^{(\tau)}$ in a distributed fashion.
The idea proposed in \cite{duchi2011dual} is to employ a consensus-based mechanism:
\begin{equation}\label{chap1:dda-cons}
z_{i}^{(t+1)}=\sum_{j \in \calN_i \cup \{ i \}}w_{ij}{z}_{j}^{(t)}+g_i^{(t)},
\end{equation}
which is followed by a local dual averaging step
\begin{equation}\label{chap1:c-dda-lda}
    {x}_{i}^{(t+1)} =\argmin_{x\in\mathcal{X}} \left\{ \alpha_t  \left\langle z_{i}^{(t+1)}, x \right\rangle + d(x)  \right\}.
\end{equation}
The resulting distributed dual averaging, however, has sub-linear convergence \cite{duchi2011dual,liu2023rate}, although a modified version achieved linear convergence in \cite{liu2022decentralized}.

\smallskip


Having explored gradient-based algorithms in the previous sections, we now shift our focus to \textit{primal-dual methods}.

\paragraph{Alternating direction method of multipliers}

Let us start by reformulating the consensus optimization problem~\eqref{eq:decentralized-optimization} as the equivalent
\begin{equation}\label{eq:decentralized-optimization-admm}
    \min_{x_i \in \mathbb{R}^d, \ \forall i} \,\, \left\{\frac{1}{n}\sum_{i=1}^n f_i(x_i)  \right\} \quad\text{s.t.} \quad x_i = y_{ij}, \,\, x_j = y_{ji}, \,\, y_{ij} = y_{ji},
\end{equation}
where the \textit{bridge} variables $y_{ij}$, $y_{ji}$, two for each edge, are introduced to decouple the consensus constraints.
We define the \textit{augmented Lagrangian} of~\eqref{eq:decentralized-optimization-admm}:
\begin{equation}
    \calL(\xx, \yy, \ww) = \frac{1}{n} \sum_{i = 1}^n f_i(x_i) + g(\yy) + \langle \ww, \AA \xx - \yy \rangle + \frac{\rho}{2} \norm{\AA \xx - \yy}^2
\end{equation}
where $\rho > 0$ is a \textit{penalty} parameter,
$\ww$ the Lagrange multipliers, $\AA$ a matrix encoding the constraints $x_i = y_{ij}$, $x_j = y_{ji}$, and $g(\yy) = 0$ if $y_{ij} = y_{ji}$ $\forall i, \, j \in \calN_i$, $g(\yy) = +\infty$ otherwise.
The alternating direction method of multipliers (ADMM) \cite{boyd2010distributed} is then defined by applying alternating minimization to the augmented Lagrangian
\begin{subequations}
\begin{align}
    \xx^{(t+1)} &= \argmin_{\xx} \calL(\xx, \yy^{(t)}, \ww^{(t)}) \\
    \yy^{(t+1)} &= \argmin_{\yy} \calL(\xx^{(t+1)}, \yy, \ww^{(t)}) \\
    \ww^{(t+1)} &= \ww^{(t)} + \rho \left( \AA \xx^{(t+1)} - \yy^{(t+1)} \right),
\end{align}
\end{subequations}
which, exploiting the distributed structure of the problem yields Algorithm~\ref{alg:admm} \cite{bastianello_asynchronous_2021}.

\begin{algorithm}[t]
\caption{Distributed ADMM}
\label{alg:admm}
\begin{algorithmic}[1]
    \Require initial conditions $y_{ij}^{(0)}$, $w_{ij}^{(0)}$, $i \in [n]$, $j \in \calN_i$, penalty $\rho > 0$.
    
    \For{$t = 0,1,\ldots$}
    \CommentState{local training}
    \For{each \textit{active} agent $i$} 
    \State compute
    \begin{small}
    \begin{equation*}
               x_i^{(t+1)} = \argmin_{x_i \in \R^d} \left\{ f_i(x_i) + \langle x_i, \sum_{j \in \calN_i} w_{ij}^{(t)} - \rho y_{ij}^{(t)} \rangle + \frac{\rho |\calN_i|}{2} \norm{x_i}^2 \right\} 
    \end{equation*}
        \end{small}
    \State transmit $x_i^{(t+1)} - w_{ij}^{(t)}$ to each neighbor $j \in \calN_i$ 
    \EndFor
    
    \CommentState{consensus}
    \For{each agent $i$ and \textit{active} neighbor $j \in \calN_i$}
    \State receive $x_j^{(t+1)} - w_{ji}^{(t)}$ and compute
        \begin{small}
    \begin{align*}
           y_{ij}^{(t+1)} &= \frac{1}{2\rho} \left( x_i^{(t+1)} - w_{ij}^{(t)} + x_j^{(t+1)} - w_{ji}^{(t)} \right), \\
        w_{ij}^{(t+1)} &= w_{ij}^{(t)} - \rho \left( x_i^{(t+1)} - y_{ij}^{(t+1)} \right)      
    \end{align*}
    \end{small}
    \EndFor
    \EndFor
\end{algorithmic}
\end{algorithm}

\paragraph{Revisiting the challenges of decentralized architectures}
The previous sections illustrated how reaching consensus in a distributed architecture poses a significant design challenge.
Additionally, distributed architectures are also subject to the challenges of \textit{communications bottlenecks}, \textit{resources heterogeneity}, and \textit{statistical heterogeneity}, discussed in Section~\ref{subsubsec:federated-opt}.

First of all, both gradient-based algorithms and dual methods require peer-to-peer communications, which might need to be quantized/compressed to reduce their burden.
Interestingly, while ADMM is robust to quantization \cite{bastianello_online_2023}, gradient tracking in general is not \cite{bin2022stability}, and requires specific design changes.
The weakness of gradient tracking descends from its reliance on (dynamic) average consensus, which is not robust to communication errors.
Similarly, average consensus is not robust to the partial participation \cite{bof_average_2017}, usually called \textit{asynchrony} in this context, resulting from resources heterogeneity -- while ADMM is \cite{bastianello_asynchronous_2021}.
The solution, therefore, is to robustify gradient tracking by replacing average consensus with a different protocol, see \textit{e.g.}, \cite{bof_multiagent_2019,tian_achieving_2020,carnevale_admmtracking_2023}.
While inherently robust to asynchrony and communication errors, ADMM (and dual averaging as well) presents a different challenge: the agents are required to solve an optimization problem to update $x_i^{(t)}$. However, given their limited (and heterogeneous) resources, the agents may be able to only compute an inexact update \cite{bastianello_online_2023}.

\subsection{Constraint-coupled problems}
The previous sections focused on the consensus optimization problem of~\eqref{eq:decentralized-optimization} which models a broad class of decentralized learning and optimization problems. This problem is also referred to as \textit{cost-coupled}, as the agents' data is combined though the sum of local loss functions.
Another class of problems with relevance in several decentralized applications is that of \textit{constraint-coupled} \cite{notarstefano_distributed_2019}, characterized by
\begin{equation} \label{coupled_constraints}
    \min_{ x_i\in\mathbb{R}^{d_i}, \forall i }  \quad \sum_{i=1}^{n}f_i(x_i) \quad \mbox{s.t.}  \quad  \sum_{i=1}^{n} g_i(x_i)\leq {0},   \quad x_i\in\mathcal{X}_i, \,\, \forall i,
\end{equation}
where the local decisions $\{ x_i \in \R^{d_i} \}_{i = 1}^n$ are coupled through the global constraint $\sum_{i=1}^{n} g_i(x_i)\leq {0}$, $g_i : \R^d \to \R^M$, and additionally are subject to the local constraints $x_i\in\mathcal{X}_i$.
Problem~\eqref{coupled_constraints} finds application in resource allocation \cite{xiao2006optimal,turan2020resilient}, distributed charging control of electric vehicles \cite{liu2017decentralized}, distributed control of multi-agent systems \cite{wang2018accelerated,fang2023distributed}, \textit{etc}.
The algorithms discussed in Sections~\ref{subsubsec:federated-opt} and~\ref{subsubsec:distributed-opt} cannot be directly applied to solve~\eqref{coupled_constraints} and tailored approaches need to be designed, discussed briefly in the following. Sections \ref{sec:private-optimization} and~\ref{sec:resilient-optimization} will further illustrate how privacy and robustness mechanisms can be tailored to this class of problems.

A widely used approach to solving~\eqref{coupled_constraints} is to leverage the fact that its dual is a cost-coupled problem \cite{notarstefano_distributed_2019}. In particular, the dual optimization problem of~\eqref{coupled_constraints} is
\begin{equation}\label{chap1:dual_lagr}
	\max_{w \geq{0}} \left\{ \sum_{i=1}^{n}\psi_i(w) \right\}, 
\end{equation}
where
$
    \psi_i(w) := \min_{x_i\in\mathcal{X}_i}\left\{  f_i(x_i)+\left \langle w, g_i(x_i) \right\rangle \right\}
$
and $w \in \R^M$ is the vector of dual variables. Clearly, \eqref{chap1:dual_lagr} is a cost-coupled problem and -- in principle -- the algorithms of Sections~\ref{subsubsec:federated-opt} and~\ref{subsubsec:distributed-opt} can now be applied to it. However, notice that the local costs $\psi_i$ are defined through a minimization problem, and may not be smooth. Different approaches can be taken to solve this issue, and we reference \cite{mateos2016distributed,falsone2017dual,notarnicola2019constraint,liang2019distributed} for some examples.

Solving the cost-coupled dual of~\eqref{coupled_constraints}, however, only guarantees asymptotic satisfaction of the constraint. This approach is thus not suitable for safety-critical systems, in which the constraints need to be verified at all times.
The alternative is to address directly a suitably reformulated version of the primal problem~\eqref{coupled_constraints}, see \cite{tan2021distributed,mestres2023distributed,wu2021new,wu2022distributed} for some examples.

%% file: paper/private_DO.tex
\section{Privacy-preserving decentralized optimization}\label{sec:private-optimization}

{In the decentralized algorithms discussed in Section~\ref{sec:preliminaries}, the agents cooperate by sharing locally trained models (either with the coordinator or their neighbors). However, sharing these models potentially exposes the agents' private data. This section will start by reviewing the prevalent cryptographic tools employed to avoid data leakage. This will be followed by a comprehensive review of privacy-preserving decentralized optimization and consensus algorithms.}


In this section, we consider the following three types of privacy attacks. Some examples are provided in Table \ref{tab:attack}.

	\begin{itemize}
	\item \textit{Honest-but-curious agents or central coordinator}:  These agents or the central coordinator follow the algorithm to perform communication and computation. However, they may record intermediate results to infer sensitive information about other agents.

	\item \textit{Colluding agents}: Certain agents may collude with the central coordinator or amongst themselves to deduce private information about a specific victim agent.
	\item \textit{Outside eavesdropper}: This attacker can intercept all shared messages during the execution of the training protocol but will not actively inject false messages or interrupt message transmissions.  
	\end{itemize}
 
\begin{table*}[t]
\centering
\caption{Examples of privacy attacks.}
\label{tab:attack}
    \begin{tabular}{@{}ccc@{}}
    \hline 
    {Attack 
    vectors}   &  Description & {Representative references}     \\ 
        \hline
    {feature/attribute inference}  & \thead{derive an individual's characteristics \\ from the intermediate computation results}  & \cite{hitaj2017deep,melis2019exploiting} \\
    membership inference  & \thead{determine whether a sample was \\ used to train the model 
    (tracing attack)} & \cite{shokri2017membership,nasr2019comprehensive} \\
    data construction  & \thead{reconstruct training samples \\and associated labels accurately}& \cite{zhu2020deep,geiping2020inverting} \\
    \hline
    \end{tabular}
\end{table*}





\subsection{Cryptographic tools}
\label{subsec:tools}

In the field of optimization and learning, key privacy-preserving techniques include differential privacy \cite{dwork2006calibrating}, homomorphic encryption \cite{rivest1978method}, and the secret-sharing protocol \cite{shamir1979share}.

\subsubsection{Differential privacy}

In recent years, differential privacy (DP) has become prevalent in cryptography and machine learning \cite{dwork2006differential,abadi2016deep}, owing to its precise mathematical formulation and computational tractability. Informally, DP ensures that the output of an algorithm remains insensitive to the presence or absence of any individual's data in the dataset, thereby protecting individuals from privacy attacks that aim to extract their personal information.

Next, we review some key concepts in DP.
Let $\mathcal{D}=\{\xi^{1},\dots, \xi^{q}  \}$ denote a dataset of size $q$ with records drawn from a universe $\mathbb{X}$ (\textit{e.g.}, $\R^d$). Two datasets $\mathcal{D}$ and $\mathcal{D}'$ are referred to as \textit{neighboring} if they are of the same size and differ in one data point. This is particularly relevant to machine learning where the optimization problem is defined over datasets, see Example \ref{em:ERM}.



\begin{definition}[Differential privacy]
\label{def_DP}
Given $\varepsilon,\delta\geq 0$, a randomized function $\mathcal{M}: \mathbb{X}^q \rightarrow \mathbb{Y}$ is said to be $(\varepsilon,\delta)$-DP, if for every pair of neighboring datasets $\mathcal{D}$, $\mathcal{D}'\in\mathbb{X}^q$ and every subset $\mathcal{O}\subseteq \mathbb{Y}$, we have
\begin{equation}
\label{eq_DP}
\begin{split}
    \mathbb{P}[\mathcal{M}(\mathcal{D})\in\mathcal{O}]\leq  e^{\varepsilon} \mathbb{P}[\mathcal{M}(\mathcal{D'})\in\mathcal{O}]+\delta.
\end{split}
\end{equation}
In addition, when $\delta=0$ in \eqref{eq_DP}, $\mathcal{M}$ is said to be $\varepsilon$-DP.
\end{definition}

Definition \ref{def_DP} highlights the indistinguishability of the datasets $\mathcal{D}$ and $\mathcal{D}'$ based on the output of the function $\mathcal{M}$. To see this, we consider the case $\delta=0$, for which it holds
\begin{equation*}
\frac{\mathbb{P}[\mathcal{M}(\mathcal{D})\in\mathcal{O}]}{\mathbb{P}[\mathcal{M}(\mathcal{D'})\in\mathcal{O}]}\leq e^{\varepsilon}.
\end{equation*}
The numerator represents the probability of observing the output $\mathcal{O}$ when the function $\mathcal{M}$ is applied to the dataset $\mathcal{D}$, while the denominator represents the probability of observing the same output $\mathcal{O}$ when $\mathcal{M}$ is applied to the neighboring dataset $\mathcal{D}'$. The requirement that the ratio of these two probabilities be bounded by $e^{\varepsilon}$ implies that the outputs of $\mathcal{M}$ on the two datasets are indistinguishable, as their probabilities of occurrence cannot differ significantly.

We present two fundamental mechanisms that are widely used to achieve DP in the field of optimization and learning: the Laplace and Gaussian mechanisms.


\begin{lemma}[Laplace mechanism]\label{Laplace_to_DP}
	Consider the Laplace mechanism for answering the query $r:\mathcal{D}\rightarrow \mathbb{R}^d$:
		\begin{equation*}
		\mathcal{\mathcal{M}}_L(\mathcal{D}) = r(\mathcal{D}) +(\nu_1,\dots,\nu_d),
	\end{equation*} 
where $\nu_i, \forall i$ are independent and have the following probability density function $ e^{-\frac{{x}}{\lambda}}/(2\lambda)$ where $\lambda>0$ is a scale parameter. The mechanism $\mathcal{M}_L$ is ($\Delta_1/\lambda,0$)-DP where $\Delta_1$ denotes the $\ell_1$-sensitivity of $r$, \textit{i.e.},
	$
	\Delta_1  = \sup_{\mathcal{D}, \mathcal{D}'}\lVert  r( \mathcal{D})-r( \mathcal{D}') \rVert_1$ where $\mathcal{D}$ and $\mathcal{D}'$ are neighboring datasets.
\end{lemma}



\begin{lemma}[Gaussian mechanism]\label{Gaussian_to_DP}
	Consider the Gaussian mechanism for answering the query $r:\mathcal{D}\rightarrow \mathbb{R}^d$:
		\begin{equation*}
		\mathcal{\mathcal{M}}_G (\mathcal{D})= r(\mathcal{D}) +\nu,
	\end{equation*} 
where $\nu\sim \mathcal{N}(0, \sigma^2 I_d)$.
The mechanism $\mathcal{M}_G$ is ($\sqrt{2\log(2/\delta)}\Delta_2/\sigma,\delta$)-DP where $\Delta_2$ denotes the $\ell_2$-sensitivity of $r$, \textit{i.e.},
	$
	\Delta_2  = \sup_{\mathcal{D}, \mathcal{D}'}\lVert  r( \mathcal{D})-r( \mathcal{D}') \rVert
	$ where $\mathcal{D}$ and $\mathcal{D}'$ are neighboring datasets.
\end{lemma}

Next, we take the projected gradient descent in \eqref{eq:distributed-pgd} applied to the ERM problem in \eqref{eq:centralized-erm} as an example to illustrate the Gaussian mechanism.

\begin{example}[Differentially private gradient descent]
Denote by $\mathcal{D}_i=\{ \xi_i^h \}_{h = 1}^{m}$ the local dataset to agent $i$. Then, the local gradient descent update in \eqref{eq:distributed-pgd-1} can be rewritten as
\begin{equation*}
    y_i^{(t+1)} = x_i^{(t)} - \alpha r^{(t)}(\mathcal{D}_i)
\end{equation*}
where $\alpha$ is the step-size and
\begin{equation*}
  r^{(t)}(\mathcal{D}_i)= \sum_{h=1}^m \nabla \ell (x_i^{(t)},\xi^h_i). 
\end{equation*}
To make the gradient query $r^{(t)}$ differentially private, we employ the Gaussian mechanism
\begin{equation*}
   \mathcal{\mathcal{M}}^{(t)}_G  = r^{(t)}(\mathcal{D}_i)+\nu,
\end{equation*}
where $\nu\sim \mathcal{N}(0, \sigma^2 I_d)$. As elaborated in Lemma \ref{Gaussian_to_DP}, the privacy level of $\mathcal{M}_G^{(t)}$ is determined by $\sigma$ and the sensitivity of $r^{(t)}(\mathcal{D}_i)$, defined by
\begin{equation*}
    \Delta_2  
    = \sup_{\xi_i\in \mathcal{D}_i, \xi'_i\in\mathcal{D}'_i}\lVert  \nabla \ell (x_i^{(t)},\xi_i) - \nabla \ell (x_i^{(t)},\xi'_i) \rVert.
\end{equation*}

\end{example}

We note that the privacy guarantees provided in Lemmas \ref{Laplace_to_DP} and \ref{Gaussian_to_DP} are for a single query, \textit{i.e.}, a single iteration of the optimization algorithm. 
The privacy leakage can accumulate when multiple queries are made to a differentially private mechanism.
Finding tighter bounds on the cumulative privacy leakage for multiple queries is an active area of research \cite{altschuler2022privacy}.
In the simplest case, the DP parameter grows linearly with respect to the number of queries \cite{dwork2014algorithmic}. Advanced composition of multiple queries and the corresponding privacy leakage have been studied in \cite{kairouz2015composition}. Tight composition bounds were also provided along with other statistical privacy definitions, such as Rényi DP \cite{mironov2017renyi}.

DP can be achieved through other mechanisms as well, \textit{e.g.}, the exponential mechanism \cite{mcsherry2007mechanism} and the median mechanism \cite{roth2010interactive}. For special scenarios where the query outputs are bounded, such as interval observer design for dynamical systems \cite{degue2020differentially}, the \textit{truncated} Laplace mechanism \cite{geng2020tight} and the \textit{truncated} Gaussian mechanism \cite{liu2018generalized} can be employed. Furthermore, while both the Laplace and Gaussian mechanisms have analytic expressions for the noise, numerically optimized additive noise mechanisms have been discussed in \cite{sommer2021learning}. Finally, we remark that the DP constraint typically induces a trade-off between privacy and utility in learning algorithms \cite{bassily2014private,kifer1private,chaudhuri2012near}, where the utility quantifies the accuracy of a trained model.



\subsubsection{Homomorphic encryption}

Homomorphic encryption (HE) is a classical encryption scheme \cite{rivest1978data,paillier1999public}.
The goal in cryptography is to take a message or data, called plaintext, and provide an encrypted version, called cyphertext, which obscures the original text. Indeed, without the correct decryption key or algorithm, it is computationally infeasible to recover the original plaintext from the ciphertext, thereby ensuring the confidentiality of the information.
A fully homomorphic encryption scheme (FHE) refers to a scheme, where \textit{any} computations on the plaintexts can be obtained by directly manipulating ciphertexts so that the privacy of the underlying plaintexts is preserved. Such property for the encryption and decryption operations is referred to as \textit{homomorphic}, which describes the preservation of the algebraic structure with respect to a certain arithmetic function.
FHE is achievable if and only if addition and multiplication operations can be homomorphically performed \cite{marcolla2022survey}. More formally, let $x\in\mathbb{Z}$ and $y\in\mathbb{Z}$ be the plaintexts. Their ciphertexts are denoted as $\text{Enc}(x)$ and $\text{Enc}(y)$, respectively, where $\text{Enc}(\cdot)$ represents the encryption operation. The encryption of $x+y$ (or $x \cdot y$) can be obtained by simply adding or multiplying the corresponding ciphertexts, \textit{i.e.},
\begin{equation}\label{eq:he_property}
    \text{Dec}(\text{Enc}(x) \,\, \triangle \,\, \text{Enc}(y) )= x \,\, \triangle \,\, y
\end{equation}
where $\triangle$ represents either the addition or multiplication operation, and $\text{Dec}(\cdot)$ is the decryption operation.
We remark that when dealing with non-integer plaintexts, quantization techniques can be employed to convert them into integer representations. However, the quantization process introduces approximation errors, and these quantization effects on the overall system performance should be properly handled.

It is important to note that the encryption procedure introduces random noise, and the error accumulates as more homomorphic operations are performed, which may eventually prevent the correct decryption of ciphertexts. Consequently, the homomorphic property holds only for a bounded number of operations. To address this issue, the \textit{bootstrapping} technique has been employed \cite{gentry2009fully}, which outputs a new ciphertext with reduced noise based on a ciphertext with high noise and a bootstrapping or refreshing key. This process enables an unlimited number of homomorphic operations to be performed. Importantly, this procedure is carried out without decrypting the input ciphertext, meaning that it is a homomorphic evaluation of the decryption procedure. Bootstrapping is typically the most sophisticated and computationally demanding component of an FHE scheme \cite{al2023demystifying}.

A cryptosystem is called \textit{additively homomorphic} if \eqref{eq:he_property} only holds for sum, \textit{e.g.}, Paillier cryptosystem \cite{paillier1999public}, and \textit{multiplicatively homomorphic} if \eqref{eq:he_property} holds for product, \textit{e.g.}, RSA cryptosystem \cite{rivest1978method}. Any additively homomorphic cryptosystems also support encrypted multiplications with (un-encrypted) constant numbers, that is, given $k\in\mathbb{N}$
\begin{equation*}
    k \cdot \text{Enc}(y) = \underbrace{\text{Enc}(y) \,\, +  \,\, \text{Enc}(y) + \cdots + \,\, \text{Enc}(y)}_{k \,\, \text{times}}.
\end{equation*}
Extensions can be made to the case with $k\in\mathbb{Z}$ and multiplications with integer matrices. Therefore, although being less powerful than FHE, additively homomorphic encryption offers a computationally more tractable approach for designing privacy-preserving optimization and control algorithms \cite{ruan2019secure,kim2022comparison}.

\begin{example}[Paillier cryptosystem and its application in privacy-preserving optimization]

The Paillier cryptosystem \cite{paillier1999public} can be summarized in three key steps:
\begin{itemize}
    \item[1)] \textit{Key generation}: 
    Pick two large prime numbers $p$ and $q$ satisfying $\text{gcd}(pq,(p-1)(q-1))=1$ where $\text{gcd}$ represents the greatest common divisor. Let $N=pq$. Choose $1\leq g\leq N^2$ randomly and compute $\lambda = \text{lcm}(p-1,q-1)$ where $\text{lcm}$ is the least common multiple. Compute $\mu$ as a modular multiplicative inverse of $L(g^{\lambda} \,\, \text{mod} \,\, N^2)$, \textit{i.e.}, $\mu \cdot L(g^{\lambda} \,\, \text{mod} \,\, N^2) \equiv 1\,\, \text{mod} \,\, N$, where $L(x)=\frac{x-1}{N}$.
    The public key is $(N,g)$. The private key is $(\lambda, \mu)$.
    
    \item[2)] \textit{Encryption}: To encrypt an (integer) message $m$:
\begin{itemize}
    \item[i)] Choose a random integer $0< x < N$,
    \item[ii)] Produce an encryption $\text{Enc}(m) = g^mx^N \,\, \text{mod} \,\, N^2$.
    \end{itemize}

 Through arithmetic operations, it can be demonstrated that the Paillier cryptosystem exhibits an additive homomorphic property, which is expressed as follows:
\begin{center}
\begin{tabular}{  m{2.5cm}  m{3.6cm} } 
  \hline
  $\text{Enc}(m_1)$ & $=g^{m_1}x_1^N \,\, \text{mod} \,\, N^2 $   \\ 
  $\text{Enc}(m_2)$ & $=g^{m_2}x_2^N \,\, \text{mod} \,\, N^2 $  \\ 
  \hline
  $\text{Enc}(m_1)\cdot \text{Enc}(m_2)$ & 
  $=g^{m_1+m_2}(x_1x_2)^N \,\, \text{mod} \,\, N^2$  \\  & = $\text{Enc}(m_1+m_2) $   \\ 
  \hline
\end{tabular}
\end{center}

    \item[3)] \textit{Decryption}: Let $c$ be the ciphertext to decrypt. The plaintext message can be exactly recovered as
\begin{equation*}
    m =  L(c^{\lambda} \,\, \text{mod}\,\, N^2)  \cdot \mu \,\,\text{mod} \,\, N
\end{equation*}
\end{itemize}

The correctness of the algorithm was proved in \cite{paillier1999public}. Thanks to its additively homomorphic property, Paillier's cryptosystem can be conveniently employed to implement the optimization algorithm. Recall the gradient descent update step given by \eqref{eq:distributed-pgd-1}:
\begin{equation*}
    y_i^{(t+1)} = x_i^{(t)} - \alpha \nabla f_i(x_i^{(t)}). 
\end{equation*}
To perform this update step in an encrypted manner, we can utilize Paillier cryptosystem as shown in \cite{aono2017privacy}:
\begin{align*}
     \text{Enc}(y_i^{(t+1)})  &= \text{Enc}(x_i^{(t)}-\alpha \nabla f_i(x_i^{(t)})) \\
     &=   \text{Enc}(x_i^{(t)})\cdot \text{Enc}(-\alpha \nabla f_i(x_i^{(t)})).
\end{align*}

\end{example}


\subsubsection{Secret sharing}

Secret sharing (SS) protocols are encryption methods for distributing a confidential message in the form of ``shares" to multiple parties. The message remains confidential {as long as the attacker can only intercept a limited number of shares}. Specifically, an $(n,k)$-threshold SS scheme consists of two sub-algorithms: \textit{Share} and \textit{Reconstruct}. The \textit{Share} algorithm takes a private message $S$ as input and outputs $n$ secret shares $\{ S_1, \dots, S_n \}$. The \textit{Reconstruct} algorithm takes $k$ secret shares $\{ S_i \}_{i\in\mathcal{I}}$ with $\lvert \mathcal{I} \rvert=k$ and $\mathcal{I}\subseteq [n]$, and recovers $S$ as the output. Each share $S_i$ is distributed to one party $i\in[n]$. The requirement is that any collusion of less than $k$ parties cannot reconstruct the private message $S$. Each party is able to perform computations on its share and generate an outcome, from which the data owner can reconstruct the computation result.

Shamir's scheme \cite{shamir1979share} is a well-known SS scheme in which the secret shares are generated by evaluating a polynomial of degree $(k-1)$ over $n$ different points. In particular, consider a polynomial of degree $(k-1)$:
\begin{equation*}
    p(\theta) = \sum_{\ell =1}^{k-1} a_\ell \theta^\ell+x
\end{equation*}
where $x$ denotes an integer secret, $a_1,\dots,a_{k-1}$ are random coefficients that are uniformly distributed in the finite field of $\{ 0,1,\dots, M-1 \}$, and $M$ is a prime number larger than $x$. By setting $\theta=1,\dots,n$, the data owner obtains $ \{ p_1^{\Pi},\dots, p_n^{\Pi} \}$, where
\begin{equation*}
    p_\theta^{\Pi} = p(\theta) \,\, \text{mod} \,\, M.
\end{equation*}
At least $k$ secret shares are required to reconstruct the secret $x$ by
\begin{equation*}
    x = \sum_{\ell=1}^k p_{\ell}^{\Pi} \prod_{\nu=0 \atop \nu\neq \ell}^k \frac{\nu}{\nu-\ell}.
\end{equation*}

SS offers a scalable, robust, and secure approach to safeguarding sensitive information. One of its key advantages is the ability to reconstruct the original secret data from a predetermined number of, not necessarily $n$, shares. This feature becomes particularly valuable in scenarios where nodes fail due to various reasons, as the secret can still be recovered from the remaining valid shares. Additionally, SS is often combined with other cryptographic techniques, such as {encryption \cite{bonawitz2017practical}}, thereby providing an additional layer of security.

Next, we use an example to illustrate how SS can be used to implement the model aggregation step, \textit{i.e.}, \eqref{eq:distributed-pgd-2-bis}, in a privacy-preserving manner.
\begin{example}[Privacy-preserving aggregation based on secret sharing \cite{emekcci2007privacy}]
Assume there are four agents, each with a secret value $x_1$, $x_2$, $x_3$, and $x_4$, respectively {-- for example, in~\eqref{eq:distributed-pgd} we have $x_i := x_i^{(t)} - \alpha \nabla f_i(x_i^{(t)})$}. They want to compute the sum of their secret values, $\sum_{i=1}^4 x_i$, without revealing their individual secrets. To achieve this, they agree on a polynomial degree $k=3$ and the number of shares $n=4$.

The protocol works as follows: Each agent $i$ constructs a random polynomial $p_i(\theta)$ of degree $3$ in the following form:
\begin{equation*}
p_i(\theta) = a_{i,3} \theta^3 + a_{i,2} \theta^2 + a_{i,1} \theta + x_i
\end{equation*}
where $a_{i,3}$, $a_{i,2}$, and $a_{i,1}$ are randomly chosen coefficients, and $x_i$ is the agent's secret value.
Each agent $i$ computes the values of their polynomial $p_i(\theta)$ for $\theta = 1, 2, 3, 4$, and shares $p_i(j)$ with agent $j$, respectively.
After receiving the shares $p_j(i)$ from all other agents $j$, each agent $i$ computes the sum of the received shares: $\sum_{j=1}^n p_j(i)$.
Each agent $i$ sends the sum of shares $\sum_{j=1}^n p_j(i)$ to all other agents.
At this point, each agent knows the values of the sum polynomial $p(\theta) = \sum_{i=1}^n p_i(\theta)$ at points $\theta = 1, 2, 3, 4$, which are:
\begin{equation*}
p(\theta) = \left(\sum_{i=1}^n a_{i,3} \right) \theta^3 + \left(\sum_{i=1}^n a_{i,2} \right) \theta^2 + \left(\sum_{i=1}^n a_{i,1} \right) \theta + \sum_{i=1}^n x_i.
\end{equation*}
Since each agent knows four points on the sum polynomial $p(\theta)$, which has degree $3$, they can use polynomial interpolation to recover the coefficients of $p(\theta)$, including the constant term $\sum_{i=1}^n x_i$, which is the desired sum of their secret values.

\end{example}


A comparison of the three cryptographic tools is presented in Table \ref{tab:comparison-cryptographic}. DP is designed to prevent data leakage from trained models that are disclosed. Encryption and secret sharing, instead, have the goal of obscuring the disclosed model itself. It is important to note that these tools are complementary and can be combined to further enhance privacy protection; see Section \ref{sec:discussion_on_privacy} for more discussions. 


\begin{table*}[t]
\centering
\caption{A comparison of main cryptographic tools.}
\label{tab:comparison-cryptographic}
    \begin{tabular}{@{}cccccc@{}}
    \hline
  {Cryptographic} &  \multirow{2}{*}{Utility} &  \multirow{2}{*}{Privacy guarantee} & Computation and  &  \multirow{2}{*}{Feature} & Representative \\
      tools &&& communication load & & references  \\
    \hline 
    DP  & inexact & \thead{statistical \\ indistinguishability} &   low & \thead{easy \\ implementation} & \cite{dwork2006differential,abadi2016deep} \\ 
    HE  & exact & \thead{computational \\ indistinguishability} & high &  \thead{operations on \\encrypted data} & \cite{marcolla2022survey} \\ 
    SS  & exact & \thead{computational \\indistinguishability } &   medium & \thead{resilient  \\to dropouts}  & \cite{shamir1979share} \\  
    \hline
    \end{tabular}
\end{table*}

{In the following sections, we review how the cryptographic tools of the previous section have been applied to prevent data leakage in decentralized optimization and learning.}

\subsection{Differential privacy-based algorithms}\label{subsec:DP_algorithms}


We categorize the existing DP-based decentralized optimization algorithms based on their communication topologies. Following this categorization, we present a discussion on privacy amplification measures for DP-based algorithms, {which leverage specific components of an algorithm, \textit{e.g.}, quantization, to enhance privacy}.

\subsubsection{Federated topology} 
{For federated learning with heterogeneous data, the authors in \cite{hu2020personalized} developed a personalized linear model training algorithm with DP. In \cite{noble2022differentially}, general models were considered. In particular, the subsampling of users’ local data was explicitly used to amplify the DP guarantee and improve utility.}
\cite{zhang2023dynamic} proposed locally differentially private algorithms for federated learning with strongly convex and composite objectives. These algorithms employed the Gaussian mechanism, with noise variance dynamically adjusted over time to enhance the trade-off between accuracy and privacy.
For optimization problems where each agent holds its own constraint and collaboratively minimizes an objective that is a function of the average of individual variables, \cite{han2016differentially} perturbed the gradients used in the projected gradient descent step performed by each agent. The algorithm featured a step-size of $\Theta(\nicefrac{1}{\sqrt{t}})$, and the utility loss was on the order of $\mathcal{O}\left(\sqrt[4]{\nicefrac{dn^2}{\varepsilon}}\right)$, where $n$ denotes the number of agents and $d$ denotes the dimension of the decision variable.


\subsubsection{Peer-to-peer topology} 
We review DP-based distributed algorithms operated over peer-to-peer topologies, dividing them based on the objective that they accomplish: consensus or optimization.
It is worth mentioning that, in the consensus literature, existing works usually aim to protect the initial states of agents, while in the distributed optimization literature, the focus is different. Some privacy-preserving distributed optimization works concentrate on protecting datasets, which is particularly relevant in machine learning problems.
Alternatively, other works in the distributed optimization domain emphasize protecting the individual cost functions of the agents.




\paragraph{Consensus} \cite{huang2012differentially} studied the private iterative consensus problem, where agents are required to converge while protecting the privacy of their initial values from honest but curious adversaries. The Laplace mechanism was used to achieve DP. Similarly, \cite{nozari2017differentially} injected Laplace noise into the consensus-seeking process to attain DP. However, they adjusted the consensus protocol so that the proposed algorithm converges almost surely to an unbiased estimate of the average of agents' initial states. 
\cite{wang2023robust} proposed a privacy-preserving distributed dynamic average consensus with noise-adding, which ensures DP.
\cite{wang2024differentially} developed a differentially private bipartite consensus algorithm operated over signed networks. The noise variance and step-size\footnote{The step-size refers to the weight assigned to neighboring information in each round.} were jointly designed to achieve asymptotically unbiased bipartite consensus. \cite{cyffers2022muffliato} introduced pairwise network differential privacy, a relaxation of local differential privacy that captures the notion that the privacy leakage from one node to another may depend on their relative positions in the graph. The authors analyzed the combination of local noise injection with gossip averaging protocols on fixed and random communication graphs, providing theoretical guarantees for the privacy and utility trade-offs achieved by these algorithms. In the context of average consensus of positive systems, \cite{wang2023differentially} proposed a novel differentially private randomized mechanism that perturbs the value using truncated Gaussian noise in a multiplicative manner. They analyzed the algorithm's performance in terms of an accuracy metric and quantified its DP guarantee. A comparison of DP-based consensus algorithms is provided in Table \ref{tab:comparison-DP-consensus}.

\begin{table*}[t]
\centering
\caption{A comparison of DP-based consensus algorithms.}
\label{tab:comparison-DP-consensus}
    \begin{tabular}{@{}ccccccc@{}}
    \hline
    \multirow{2}{*}{ [Ref.]}   & \multirow{2}{*}{Problem}  & Perturbed & {Noise  \& } & \multirow{2}{*}{Step-size$^\dagger$} & Composition$^\ddagger$ &
\multirow{2}{*}{Utility loss}   \\
 & &  term &  variance change & & range &  \\
    \hline
            \cite{huang2012differentially}  & \thead{average consensus} & states &  \thead{Laplace, \\decreasing} & constant  &  $\infty$& \thead{ $\mathcal{O}(\nicefrac{1}{\sqrt{n}\varepsilon})$ in 
     accuracy} \\ 
        \cite{nozari2017differentially}  & \thead{average consensus} & states &  \thead{Laplace,\\decreasing} & constant  &  $\infty$& \thead{exact expected value \\ of convergence,\\ $\mathcal{O}(\nicefrac{1}{\sqrt{n}\varepsilon})$
       in 
        accuracy} \\ 
        \cite{wang2023robust}  & \thead{dynamic \\
        average consensus} & states &  \thead{Laplace, \\increasing } & decreasing  &  $\infty$& -  \\  
        \cite{wang2024differentially} &  \thead{bipartite consensus} & states &  \thead{Laplace,\\increasing} & decreasing  &  $\infty$ & -  \\  
        \cite{wang2023differentially}  & \thead{positive consensus} & states &  \thead{multiplicative \\ Gaussian,\\ constant} & decreasing  &  $T$ & -  \\  
        \thead{\cite{cyffers2022muffliato} \\ (Algs. 1-2)}&\thead{gossip averaging} & states &  \thead{Gaussian\\ constant} & constant  &  $T$ & \thead{$\mathcal{O}(\nicefrac{1}{n^2 \varepsilon})$ in \\ consensus error } \\ 
      \hline
    \end{tabular}\\
    {\footnotesize $^\dagger$ The step-size refers to the weight assigned to neighboring information in each round.\\
$^\ddagger$ The composition range indicates the number of iterations considered when calculating privacy loss.}
\end{table*}

\paragraph{Distributed optimization} For distributed optimization problems, \cite{huang2015differentially} proposed a differentially private distributed gradient descent algorithm that perturbs the local output with Laplace noise. The algorithm employed a linearly decaying step-size, ensuring that the sensitivity also decreases linearly. The prescribed differential privacy parameter $\varepsilon$ was decomposed into a sequence $\{\varepsilon_t\}_{t\geq 1}$ such that $\sum_{\tau=1}^\infty \varepsilon_\tau =\varepsilon$, allowing each iteration $t$ to be $\varepsilon_t$-DP.
However, this choice of a linearly decaying step-size significantly slowed down the convergence rate and led to a utility loss of order $\mathcal{O}(\nicefrac{d}{\varepsilon^2})$, where $d$ represents the dimension of the decision variable.
Along this line of research, several works extended differentially private distributed optimization algorithms to handle time-varying objective functions \cite{zhu2018differentially,xiong2020privacy,han2021differentially}.

\cite{ding2021differentially} incorporated a gradient-tracking mechanism into the differentially private distributed optimization algorithm, enabling it to achieve a linear convergence rate. 
\cite{chen2023differentially} developed a privacy-preserving scheme based on the robust gradient-tracking distributed optimization algorithm introduced in \cite{pu2020robust}. The exchanged variable was updated with carefully calibrated Laplace noise to achieve DP. Moreover, \cite{chen2023differentially} employed the SS technique, where each individual state was split into two shares. Only the share whose update does not directly depend on the local gradient was exchanged with neighbors, thereby enhancing privacy protection.
Other existing works based on SS will be reviewed later in Section \ref{subsec:SS}. 
Inspired also by \cite{pu2020robust}, \cite{xuan2023gradient} developed a privacy-preserving distributed algorithm that supports both constant and decreasing step-sizes. To attenuate the noise effect while ensuring DP, the authors in \cite{vlaski2021graph} constructed topology-aware noise. With this method, each agent perturbed the messages to its neighbors (including itself) with different perturbations whose weighted sum was zero. \cite{huang2024differential} presented an impossibility result regarding the simultaneous achievement of DP and exact accuracy in distributed optimization.
The authors demonstrated that $(\varepsilon,0)$-DP cannot be achieved by the Laplace mechanism when employing diminishing step-sizes. Furthermore, they quantified the accuracy loss and DP parameters when utilizing linearly decaying step-sizes and noise variances. {In \cite{kalra2023decentralized}, a novel algorithm was developed where each agent maintains two models at each step: a private model and a proxy model. Communication with neighboring agents involved only the proxy model, which was trained with DP noise. The training losses included a KL divergence loss between the private and proxy models, allowing the private model to benefit from the proxy model’s updates.}

{Apart from perturbing the transmitted information, there are also works focusing on perturbing cost functions. In contrast to cases involving private data samples, this setup considers the cost as the private information. Therefore, to define DP according to Definition \ref{def_DP}, sets of cost functions are considered. In this context, \cite{nozari2016differentially} demonstrated the impossibility of achieving DP by perturbing the inter-agent messages with noise when the underlying noise-free dynamics are asymptotically stable. Motivated by this limitation, this work established a general framework for handling functional data, decomposing the objective functions into an infinite sequence of coefficients corresponding to the elements of an orthogonal basis in a separable Hilbert space and injecting noise into the infinite coefficient sequence. DP can be achieved by solving the distributed optimization problem defined by the perturbed local costs.
}

For composite empirical risk minimization problems, the ADMM and dual averaging were used to design distributed algorithms with DP in \cite{zhang2016dynamic,zhang2018improving} and \cite{liu2023privacy}, respectively. In particular, the authors in \cite{liu2023privacy} incorporated an agent subsampling procedure, {\textit{i.e.}, partial participation,} to enhance the privacy-accuracy trade-off in distributed optimization. For distributed stochastic optimization with convex or non-convex costs, \cite{wang2022quantization} employed ternary quantization for the communication among agents and proved {that the quantized distributed optimization algorithm preserves $(0,\delta)$-DP per iteration}. \cite{wang2023decentralizedNC} considered non-convex problems and developed a distributed optimization algorithm with DP that avoids convergence to local maxima and saddle points. \cite{wang2023tailoring} proposed two gradient-based algorithms for differentially private distributed optimization. These algorithms can guarantee a finite privacy budget, {\textit{i.e.}, $\varepsilon$ in Definition \ref{def_DP}}, even when the number of iterations approaches infinity, under certain conditions. A comparison of DP-based distributed optimization algorithms is given in Table \ref{tab:comparison-DP}.

\begin{table*}[t]
 \centering
\caption{A comparison of DP-based distributed optimization algorithms.}
\label{tab:comparison-DP}
    \begin{tabular}{@{}ccccccc@{}}
    \hline
    \multirow{2}{*}{ [Ref.]}   & {Private}  & Perturbed & {Privacy-preserving  } & \multirow{2}{*}{Step-size} & Composition &
\multirow{2}{*}{Utility loss$^\ddagger$}   \\
 & information &  term &  mechanism & & range &  \\
    \hline
        \thead{\cite{cyffers2022muffliato} \\ (Alg. 3)}  & {datasets} & variables &  \thead{Gaussian with\\ constant VAR$^\dagger$} & constant  &  $T$ & \thead{$\mathcal{O}(\nicefrac{d}{\mu n^2 \varepsilon})$ for S.C.$^*$ } \\ 
     {\cite{huang2015differentially} }  & {costs} & variables &  \thead{Laplace with \\ decreasing VAR} & decreasing  &  $\infty$ & \thead{$\mathcal{O}(\nicefrac{d}{ \varepsilon^2})$ for S.C. } \\ 
     {\cite{ding2021differentially} }  & {costs} & variables &  \thead{Laplace with \\ decreasing VAR} & constant  &  $\infty$ & \thead{$\mathcal{O}(\nicefrac{d}{ \varepsilon^2})$ for S.C. } \\
          {\cite{chen2023differentially} }  & costs & \thead{decomposed \\ variables} &  \thead{Laplace with \\ constant VAR} & constant  &  $T$ & - \\
     {\cite{xuan2023gradient} }  & costs & { variables} &  \thead{Laplace with \\ constant VAR} & \thead{constant or\\ decreasing}  &  $\infty$ & - \\
      {\cite{vlaski2021graph} }  & costs & variables &  \thead{Laplace with \\ constant VAR} & constant  &  $T$ & - \\
      {\cite{huang2024differential} }  & costs & variables &  \thead{Laplace with \\ decreasing VAR} & summable  &  $\infty$ & - \\
      {\cite{wang2022quantization} }  & gradients & variables &  \thead{ternary \\ quantization} & decreasing &  $1$ & - \\
      {\cite{wang2023decentralizedNC} }  & datasets & gradients &  \thead{Gaussian with \\ constant VAR}  & decreasing &  $1$ & - \\
      {\cite{wang2023tailoring} }  & costs & variables &  \thead{Laplace with \\ increasing VAR}  & decreasing &  $\infty$ & - \\
    \cite{liu2023privacy}  &  {datasets} &   gradients & \thead{Gaussian with \\ constant VAR} &  decreasing & $T$ & \thead{ $\mathcal{O}(\nicefrac{\sqrt{d\iota } }{\varepsilon})$ for C.\\ $\mathcal{O}(\nicefrac{{d} \iota^2 }{\varepsilon^2})$ for S.C.} \\
    {\cite{nozari2016differentially}} &{{costs}} & { costs} &  \thead{Laplace with \\ decreasing VAR} & {\thead{square \\ -summable}}  &  {$\infty$} & \thead{$\mathcal{O}(\nicefrac{n}{\varepsilon})$ for S.C. }\\
      \hline
    \end{tabular}
    \\
{\footnotesize 
$^\dagger$ VAR represents variance. \\
$^\ddagger$ For utility loss, the abbreviations C. and S.C. refer to convex and strongly convex functions, respectively. \\
$^*$ The variable $\iota$ represents the rate for agent sampling \cite{liu2023privacy}.}
\end{table*}

\paragraph{Resource allocation} For a class of resource allocation problems with equality constraints, \cite{chen2021distributed} developed a privacy-preserving distributed algorithm based on the Laplace mechanism. Considering a similar setup, \cite{ding2021differentiallyra} proposed a differentially private deviation tracking algorithm. The authors established the linear convergence of their algorithm under suitable assumptions. 

\paragraph{Game}
Inspired by~\cite{huang2015differentially}, \cite{ye2021differentially} perturbed the transmitted messages in distributed Nash equilibrium seeking with linearly decaying step-size and Laplacian noise. The authors in \cite{wang2022differentially} developed a distributed algorithm with DP for stochastic aggregative games. However, $(\varepsilon,\delta)$-DP was proved only for each iteration. \cite{chen2023differentiallygame} generalized function perturbation to the game setting and proposed a Laplace linear quadratic functional perturbation algorithm. This mechanism maintains the concavity property of the perturbed payoff functions and thus ensures the existence of Nash equilibrium of the perturbed game. Recently, \cite{wang2024differentiallygame} developed a fully distributed differentially private algorithm to achieve both rigorous DP and guaranteed computation accuracy of the Nash equilibrium.

\paragraph{Privacy amplification for DP-based algorithms}
The privacy guarantees offered by DP-based algorithms that rely on noise injection can be further enhanced through the incorporation of other (randomized) mechanisms. For instance, the synergistic effect of combining noise injection with compressed communication has been explored in \cite{yan2024killing, lang2023joint, chaudhuri2022privacy, amiri2021compressive}. Another direction is the integration with agent sampling procedures, as proposed in \cite{liu2023privacy}. By randomly selecting a subset of agents to participate in each iteration, this approach can effectively amplify the DP guarantee. Furthermore, the local training scheme (see Section~\ref{subsubsec:federated-opt}), a common component in federated learning settings, has been shown to improve privacy guarantees, as discussed in \cite{bastianello2024enhancing}.






\subsection{Homomorphic encryption-based algorithms}
In this section, we present an overview of privacy-preserving decentralized algorithms based on HE.

\subsubsection{Federated topology}
For quadratic optimization problems, where the linear term in the objective and the constant in the constraints are private, \cite{shoukry2016privacy} designed a privacy-preserving cloud-based optimization algorithm based on additively homomorphic encryption. A comprehensive theoretical analysis of this algorithm was provided later by \cite{alexandru2020cloud}. In the presence of a coordinator, \cite{lu2018privacy} proposed two privacy-preserving optimization algorithms with private and public keys, respectively, ensuring privacy security against input-output inference for quadratic functions. Considering a similar setup, \cite{huo2021privacy} developed a privacy-preserving optimization algorithm for problems with coupled costs and constraints.

HE has been actively utilized to protect privacy in machine learning over federated topologies. In \cite{aono2017privacy}, the authors employed additive homomorphic encryption for exchanging models and gradients between the coordinator and agents. This approach enables privacy preservation while retaining the training accuracy. The authors in \cite{zhang2020batchcrypt} proposed to encrypt a batch of quantized gradients from individual agents to boost computation speed and communication efficiency for federated learning with HE.

\subsubsection{Peer-to-peer topology} 
Early attempts for multi-agent consensus on encrypted values were reported in \cite{lazzeretti2014secure,freris2016distributed}. Specifically, \cite{freris2016distributed} proposed a distributed gossip algorithm that operates on encrypted values consistent with a cryptosystem initiated by a trusted third-party. An encrypted consensus scheme enabling individual agents to recover the decrypted consensus value in a fully distributed manner was later introduced in \cite{ruan2019secure}. For directed networks, \cite{gao2018privacy} presented a variant, while \cite{hadjicostis2020privacy} developed a fundamental consensus scheme on integer values and leveraged it to design a homomorphically encrypted consensus protocol. 
\cite{hung2023novel} addressed the scenario where all immediate neighbors are untrusted, potentially compromising the initial states of agents even with HE-based consensus protocol in \cite{ruan2019secure}. To mitigate this risk, they proposed the use of the Paillier cryptosystem to replace the initial states with virtual ones. A specific application of privacy-preserving consensus in directed networks to the economic dispatch problem in power grids was reported in \cite{yan2021distributed}. 

\cite{zhang2018admm} proposed a privacy-preserving distributed optimization algorithm based on {ADMM} and an additively homomorphic encryption scheme. This work was extended to constrained problems in \cite{zhang2018enabling}, where the privacy-preserving algorithm was based on the distributed projected gradient descent \cite{nedic2010constrained}. The authors in \cite{zhou2022private} adopted the Paillier encryption scheme to generate zero-sum perturbations for local objectives, and developed a privacy-preserving distributed algorithm with DP guarantee. \cite{binfet2023towards} proposed a privacy-preserving distributed ADMM algorithm for constrained quadratic problems, where a special key switching functionality was employed to enable secure computations. The authors also discussed the application of their proposed algorithm to the formation control problem.

To provide a comprehensive overview, a detailed comparison of HE-based distributed algorithms is presented in Table \ref{tab:comparison-HE}.

\begin{table*}[t]
\centering
\caption{A comparison of HE-based algorithms.}
\label{tab:comparison-HE}
    \begin{tabular}{@{}cccccc@{}}
    \hline
     \multirow{2}{*}{ [Ref.]} &  {Peer-to-peer} &  \multirow{2}{*}{ Problem}  &   {Private} &   \multicolumn{2}{c}{HE properties}  \\
     \cline{5-6} 
     & topology &  & information &  Homomorphism  & Key for agents     \\
    \hline
    \cite{shoukry2016privacy} & \xmark & \thead{cloud-based QP} & problem parameters &  additive &  common  \\
   \thead{ \cite{lu2018privacy} \\ (Alg. 1)} & \xmark & \thead{cloud-based \\
    constrained optimization} & variables and constraints &  full &  common \\
       \thead{\cite{lu2018privacy} \\ (Alg. 2)} & \xmark & \thead{ cloud-based optimization  \\
    for affine objectives} & variables and constraints &  additive &  individual \\
    \cite{huo2021privacy} & \xmark & \thead{ cloud-based \\ constrained optimization} & primal and dual variables &  additive &  common \\
    \cite{aono2017privacy} & \xmark & \thead{ federated learning} & gradients &  additive &  common \\
    \cite{zhang2020batchcrypt}  & \xmark & \thead{ federated learning} & quantized gradients &  additive &  common \\
    \cite{lazzeretti2014secure}  & \cmark & \thead{consensus} &  variables &  additive &  individual \\
    \cite{freris2016distributed}  & \cmark & \thead{consensus} &  variables &  additive &  common \\
    \cite{ruan2019secure}  & \cmark & \thead{consensus} & initial variables &  additive &  individual \\
        \cite{gao2018privacy}  & \cmark & \thead{consensus over \\ directed networks} & initial variables &  additive &  individual \\
  \cite{yan2021distributed} & \cmark & \thead{economic dispatch} & exchanged variables &  additive &  individual \\
\cite{zhang2018admm} & \cmark & \thead{distributed optimization} & variables and function &  additive &  individual \\
\cite{zhang2018enabling} & \cmark & \thead{distributed \\ constrained optimization} & variables and function &  additive &  individual \\
\cite{binfet2023towards} & \cmark & \thead{distributed constrained QP \\ and cooperative control} & variables and control objective &  additive &  individual \\
          \hline
    \end{tabular}
\end{table*}


\subsection{Secret sharing-based algorithms}
\label{subsec:SS}

Next, we review {SS-based} privacy-preserving decentralized algorithms by categorizing them based on their underlying communication topologies.

\subsubsection{Federated topology} \cite{bonawitz2017practical} proposed a privacy-preserving aggregation method for federated learning, combining masking and the $(n,p)$-threshold SS scheme. Specifically, each node adds a pairwise additive mask to its state for obfuscation, while ensuring that the sum of the original states is preserved. The masks are generated randomly using a shared seed, which is protected by the $(n,p)$-threshold SS scheme. The incorporation of the $(n,p)$-threshold SS scheme serves two purposes: first, it protects the pairwise seed used to generate the masks, and second, it provides resilience against node failures. \cite{kadhe2020fastsecagg} developed a multi-secret sharing scheme that could simultaneously share multiple secrets based on the finite field Fourier transform \cite{pollard1971fast}, easing the computation cost in massive-scale federated learning. 

\subsubsection{Peer-to-peer topology} For general peer-to-peer communication topologies, SS schemes have been successfully employed in designing privacy-preserving algorithms for average consensus and distributed optimization.

\paragraph{Consensus} \cite{wang2019privacy} proposed a privacy-preserving average consensus approach where each agent splits its state into two shares and transmits only one of them to its neighbors. The key to preserving average consensus lies in locally updating the untransmitted share in accordance with an augmented graph topology. Extensions to dynamic average consensus and push-sum consensus can be found in \cite{zhang2022privacy} and \cite{chen2023privacy}, respectively. Similar methods that mask the true state by adding deterministic offsets were discussed in \cite{altafini2019dynamical, gupta2017privacy, manitara2013privacy}. A more comprehensive study on SS-based private consensus was reported in \cite{zhang2021network}, where polynomials were used to generate shares and conditions on the privacy degree for achieving average consensus were investigated. With a common privacy degree lower than the number of neighbors, the consensus protocol exhibits resilience to node failures, a feature unique to this approach.

\paragraph{Distributed optimization} In the realm of general distributed optimization, the authors of \cite{gade2016private} proposed a novel approach to obfuscate individual objective functions by adding deliberately designed affine functions, thereby preserving the global objective function unchanged. For this algorithm, a formal statistical guarantee was subsequently provided in \cite{gupta2020preserving}. Alternatively, \cite{zhang2018privacy} considered a more general decomposition of local objective functions and enforced equivalence with the original problem by introducing consensus constraints. In the context of distributed optimization with applications to energy resource control, \cite{huo2023privacy} combined SS and the projected gradient method to design a privacy-preserving algorithm, demonstrating the versatility of these techniques across diverse domains.


To summarize, a comparison of SS-based algorithms is provided in Table \ref{tab:comparison-SS}.

\begin{table*}[t]
\centering
\caption{A comparison of SS-based algorithms.}
\label{tab:comparison-SS}
    \begin{tabular}{@{}cccccccc@{}}
    \hline
     \multirow{2}{*}{ [Ref.]} &  {Peer-to-peer} &  \multirow{2}{*}{ Problem} &  \multicolumn{2}{c}{SS properties}  &   {Resilience to }\\
     \cline{4-5} 
     & topology & & Number of shares  & Construction   &  dropouts &  \\
    \hline
    \cite{bonawitz2017practical} & \xmark & federated learning & $n\geq 2$ &  polynomial &  \cmark  \\
    \cite{kadhe2020fastsecagg} & \xmark & federated learning & $n\geq 2$ &  fast Fourier transform &  \cmark  \\
        \cite{manitara2013privacy} &  \cmark & average consensus & $\lvert \mathcal{N}_i \rvert$  & \thead{zero-sum \\ deterministic offsets} & \xmark  \\ 
    \cite{wang2019privacy} &  \cmark & average consensus & $2$  & state decomposition & \xmark  \\  
    \cite{zhang2022privacy} &  \cmark & dynamic consensus & $2$  & state decomposition & \xmark  \\  
    \cite{chen2023privacy} & \cmark & directed consensus & $2$  & state decomposition & \xmark  \\
        \cite{zhang2021network} &  \cmark & average consensus & $n\geq 2$  & polynomial & \cmark  \\ 
        \cite{gade2016private} &  \cmark & distributed optimization & $\lvert \mathcal{N}_i \rvert$  & \thead{zero-sum deterministic \\ objective perturbation} & \xmark \\
        \cite{zhang2018privacy}  &  \cmark & distributed optimization & $2$  & objective decomposition & \xmark \\
    \cite{chen2023differentially}&  \cmark & distributed optimization & $2$  & state decomposition & \xmark \\
    \cite{huo2023privacy}&  \xmark & distributed optimization & $n\geq 2$  & polynomial & \xmark \\
        \hline
    \end{tabular}
\end{table*}

\subsection{Discussions}
\label{sec:discussion_on_privacy}
In this section, we discuss the fundamental trade-off among privacy, accuracy, and efficiency of privacy-preserving decentralized algorithms.
Then, we review a few notable works that leverage the combination of multiple cryptographic techniques and also that protect privacy without relying on any of these three cryptographic tools.

\paragraph{Privacy-accuracy-efficiency trade-off} 
There exists a trade-off among privacy, accuracy, and efficiency when leveraging different cryptographic techniques for privacy-preserving decentralized algorithm design. DP-based algorithms offer rigorous privacy guarantees and are computationally and communication-wise efficient. However, they inherently degrade the accuracy of the decision or model, as there is a fundamental trade-off between DP and accuracy \cite{bassily2014private}. Conversely, HE and SS-based algorithms can preserve accuracy while providing computational indistinguishability as the privacy guarantee. Nevertheless, decentralized algorithms with (fully) HE incur significant additional computational overhead. SS-based algorithms, on the other hand, typically require agreement on the degree of polynomials and the number of shares, which can lead to high communication costs in practical scenarios.

\begin{figure}[t]
\centering
    \includegraphics[scale=0.63]{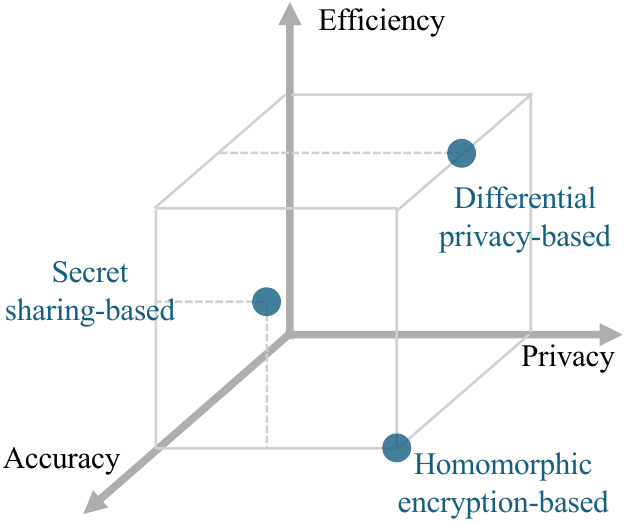}
\caption{An illustration of the privacy-accuracy-efficiency trade-off.}
\label{fig:privacy_tradeoff}
\end{figure}
	
\paragraph{Synthesis of cryptographic tools in designing distributed algorithms} The cryptographic tools introduced in Section \ref{subsec:tools} are orthogonal, and they can be combined to strengthen the privacy protection. Indeed, research studies that employ multiple cryptographic tools were reported in \cite{zhou2022private} and \cite{chen2023differentially}. In particular, \cite{zhou2022private} employed the Paillier cryptosystem to secretly construct zero-sum functional perturbations for privacy-preserving distributed optimization. This approach enables the overall algorithm to achieve DP with guarantees. As discussed in Section \ref{subsec:DP_algorithms}, the work by \cite{chen2023differentially} presented an approach that seamlessly integrated SS and DP techniques for enhancing privacy protection.

\paragraph{Distributed algorithms with other privacy-preserving mechanisms} Several notable privacy-preserving distributed algorithms used other measures to prevent privacy leakage. For example, \cite{mo2016privacy} proposed a privacy-preserving average consensus protocol that involves adding random noise to the states. The noise is generated by taking the difference between two realizations of a Gaussian distribution with linearly decaying weights at consecutive time instants. \cite{wang2023decentralized} designed two distributed optimization algorithms, one with a constant step-size and another with a diminishing step-size. They derived theoretical conditions under which a certain degree of privacy could be achieved. \cite{li2020privacy} developed a novel and general subspace perturbation method for distributed optimization. This work maintains a certain level of privacy by bounding the local information leakage, as measured by mutual information. It ensures exact convergence to the optimum without compromising privacy.

%% file: paper/robust_DO.tex
\section{Resilient decentralized optimization}\label{sec:resilient-optimization}

The goal of designing resilient decentralized algorithms is to prevent significant accuracy degradation of the decision or model in the presence of any attacks on the algorithm. The possible threat models depend on the architecture of the network. In this section, we begin by presenting the attack model and introducing some notations. Then, we discuss how the objectives in decentralized optimization and consensus-seeking problems need to be reformulated when attackers are present. This is followed by resilient aggregation and consensus mechanisms, which serve as the key module to achieve resilience in federated and distributed algorithms, respectively. 
Next, we review some metrics of resilience: aggregation resilience, graph robustness, and cost redundancy.
Following this, we provide an overview and a detailed comparison of resilient decentralized optimization algorithms in the literature.

Attackers in practical scenarios may have different objectives, such as degrading the model's accuracy or manipulating the collaborative decision (\textit{e.g.} biasing it). To comprehensively address these threats, we consider two general attack models: the Byzantine attack model \cite{lamport1983weak} and the malicious attack model. 
In the Byzantine attack model, only the honest agents correctly follow the algorithm, while the faulty agents may exhibit arbitrary and potentially \textit{different} behaviors to different neighbors. On the other hand, in the malicious attack model, the faulty agents are restricted to send the \textit{same} message to all of their neighbors, although this message may be arbitrary and differ from the intended algorithm.
In both models, the faulty agents could collude among themselves to decide on deceptive values to be communicated, further complicating the mitigation of such attacks. 
We highlight that a minimal set of requirements are imposed in these two models to constrain the adversaries.
Therefore, such attack models have been widely deployed in consensus \cite{dolev1986reaching,kieckhafer1992fault,LeBlanc2013}, federated learning \cite{feng2014distributed}, and distributed optimization \cite{sundaram2018distributed}, to ensure broad applicability of the established results.

We denote by $\mathcal{F}$ the set of faulty agents, and by $\mathcal{H}$ the collection of honest agents. It is clear that $\mathcal{H} \cap \mathcal{F} = \emptyset$ and $\mathcal{H} \cup \mathcal{F} = \mathcal{V}$, where $\mathcal{V}$ represents the set of all agents. As mentioned above, the attack model depends on the network architecture, \textit{i.e.}, federated or peer-to-peer topology; thus we distinguish the following models\footnote{The $n_f$-total attack model has been used also in the peer-to-peer scenario to simplify analysis.}:
\begin{itemize}
	\item \textit{$n_f$-total attack}: There are at most $n_f$ faulty agents in the network. That is, $|\mathcal{F}|\leq  n_f$.
	\item \textit{$n_f$-local attack}: There are at most $n_f$ faulty agents in the in-neighborhood of any agent. That is, $|\mathcal{F}\cap\mathcal{N}_i^-|\leq n_f$, for any agent $i\in\mathcal{V}$.
\end{itemize}






\subsection{Objective in the presence of attackers}\label{subsec:objective_shift}

In the presence of adversarial agents in multi-agent systems, the goals for honest agents need to be adjusted.

\paragraph{Consensus} In the multi-agent consensus problem, 
agreeing on the mean of the initial states from all agents, including the adversarial ones, is neither feasible nor sensible. Therefore, in resilient distributed consensus, the goal for the honest agents is to reach a consensus value within the convex hull formed by the initial states of the regular (non-adversarial) agents. Formally, the goal of the resilient consensus is as follows. {Design a decentralized protocol which outputs $x_i^{(t)}$, $i \in \calH$, $t \in \N$ starting from $x_i^{(0)}$, and such that:}
\begin{itemize}
	\item \textit{Safety}: Let $X^{(0)}= \{ x_1^{(0)},\dots, x_{\lvert \mathcal{H} \rvert}^{(0)}\} \subset \mathbb{R}^d$ be the set of initial states of honest agents, then at each time $t$, and for any honest agent $i$, its value $x_i^{(t)}$ should be in the convex hull of $X^{(0)}$;
 	\item \textit{Agreement}:
 As $t$ goes to infinity, it holds for any honest agent $i$ that $\lim_{t\to \infty}x_i^{(t)} = \bar{x}$ for some $\bar{x}\in\mathbb{R}^d$.
\end{itemize}
{Notice that if the safety objective is guaranteed, then $\bar{x}$ belongs to the convex hull of $X^{(0)}$.}

\paragraph{Decentralized optimization over federated or peer-to-peer topology} For the finite-sum optimization problem \eqref{eq:centralized-optimization}, solving the original problem in the presence of adversarial agents is generally not possible. Instead, a more reasonable goal is to approximate a minimizer of the following cost function:
\begin{equation}\label{eq:problem_DO_honest}
f_{\mathcal{H}}(x) = \frac{1}{|\mathcal{H}|} \sum_{i\in\mathcal{H}} f_i(x).
\end{equation}
This objective function considers only the cost functions of the honest agents, effectively excluding the adversarial agents' contributions. 

\paragraph{Resource allocation} Similarly, for the resource allocation problem in \eqref{coupled_constraints}, the problem for honest agents becomes
\begin{equation*}
\begin{split}
\min_{ x_i\in\mathbb{R}^{d_i}, \forall i \in\mathcal{H} } \ \sum_{i\in \mathcal{H}}f_i(x_i) \ \mbox{s.t.}  \ \sum_{i\in \mathcal{H}} g_i(x_i)\leq {0}   \quad x_i\in\mathcal{X}_i, \,\, \forall i\in \mathcal{H}.
\end{split}
\end{equation*}
In this new problem formulation, the cost functions and coupling constraints are modified to only account for contributions from honest agents.




Throughout Section \ref{sec:resilient-optimization}, we assume the above shifted objectives as default for resilient decentralized optimization and consensus. We note that solving the modified problems remains challenging when the identities of Byzantine agents are unknown. Consequently, existing works have typically focused on developing approximate solution methods rather than seeking exact solutions. It is noteworthy that a few recent resilient algorithms incorporate mechanisms to proactively identify attackers based on the information received during each communication round, such as reputation scoring \cite{ramos2023discrete}. As a result, these algorithms can potentially achieve improved performance.


\subsection{Resilient aggregation and consensus}\label{subsec:resilient-aggregation}




Averaging is a major step in decentralized optimization and learning algorithms over both federated and peer-to-peer topologies. To achieve resilience against Byzantine agents, standard averaging can be replaced by resilient aggregation and consensus rules. This section begins by introducing a set of common resilient aggregation rules applicable to federated topologies, followed by several useful pre-aggregation techniques. Subsequently, it reviews resilient consensus protocols over peer-to-peer topologies, emphasizing how the safety condition can be guaranteed.
Table \ref{tab:comparison-aggregation} presents a comparison of resilient aggregation and consensus rules.
Finally, it presents various metrics of resilience, including aggregation resilience expressed as a concentration inequality, as well as the notions of graph robustness and cost redundancy.

\begin{table*}[t]
\centering
\caption{A comparison of resilient aggregation and consensus rules.}
\label{tab:comparison-aggregation}
    \begin{tabular}{@{}ccccc@{}}
        \hline
     \multirow{2}{*}{Aggregation rules } &  \multirow{2}{*}{Dimension} & Safety & \multirow{2}{*}{Complexity} &  {Number of}  \\
     & &guarantee & & values removed  \\
    \hline
    coordinate-wise median  & $\geq 1$ & \xmark & $\mathcal{O}(nd)$ &   $0$  \\ 
    coordinate-wise trimmed mean  & $\geq 1$ &\xmark & $\mathcal{O}(dn\log(n))$ &   $2n_f$  \\ 
    Krum  & $\geq 1$ &\xmark & $\mathcal{O}(n^2(d+\log(n))$ &   $n_f$  \\ 
    geometric median  & $\geq 1$ & \xmark & - &   $0$  \\ 
        centered clipping  & $\geq 1$ & \xmark & $\mathcal{O}(nd)$ per step &   $0$  \\ 
    mean subsequence reduced  & $1$ & \cmark &$\mathcal{O}(n\log(n))$ &   up to $2n_f$ \\ 
    Tverberg partition-based  & $\geq 1$ & \cmark & NP &   -  \\ 
    centerpoint-based   & $\geq 1$ & \cmark & \thead{$\mathcal{O}(n)$ for $n=2$ \\ $\mathcal{O}(n^2)$ for $n=3$ }  &   - \\ 
    \hline
    \end{tabular}
\end{table*}

\subsubsection{Resilient aggregation over federated topology} 

We begin with the basic aggregation rules and then introduce two pre-aggregation techniques.


\paragraph{Coordinate-wise median (CWMed) \cite{YCKB18}} Given a set of $n$ vectors $X= \left\{x_i \in \mathbb{R}^d\right\}_{i=1}^n$, 
the CWMed of $x_1,\dots, x_n$, denoted by $\text{CWMed}(X)$, is defined as
\begin{equation*}
    \text{CWMed}(X) = \argmin_{x\in\mathbb{R}^d} \sum_{i=1}^n \lVert x - x_i \rVert_1.
\end{equation*}

\paragraph{Coordinate-wise trimmed mean (CWTM) \cite{YCKB18}}  Given a set of $n$ vectors $X= \left\{x_i \in \mathbb{R}^d\right\}_{i=1}^n$, and the largest number of faulty nodes $n_f<\nicefrac{n}{2}$. Denote by $[x_i]_k$ the $k$-th coordinate of $x_i$, and by $\tau_k$ the permutation on $[n]$ that sorts the $k$-th coordinate of $x_i, \forall i$ in a non-decreasing order, \textit{i.e.}, $[x_{\tau_k(1)}]_k \leq [x_{\tau_k(2)}]_k \leq \dots \leq [x_{\tau_k(n)}]_k$. The CWTM of $x_1,\dots, x_n$, denoted by $\text{CWTM}(X)$, is a vector in $\mathbb{R}^d$ whose $k$-th coordinate is
\begin{equation*}
    \left[ \text{CWTM}(X) \right]_k = \frac{1}{n-2n_f} \sum_{j\in[n_f+1,n-n_f]} \left[x_{\tau_k(j)}\right]_k.
\end{equation*}

\paragraph{Krum \cite{NIPS2017_f4b9ec30}} Given a set of $n$ vectors $X= \left\{x_i \in \mathbb{R}^d\right\}_{i=1}^n$, and the largest number of faulty nodes $n_f<n$. The output of Krum is a vector $x\in X$ that is closest to its neighbors upon pruning the $n_f$ farthest ones. Let $C_i$ be the set of $n-n_f-1$ closest vectors to $x_i$. Then, Krum outputs the vector that has the smallest sum of distances to its $n-n_f$ closest neighbors, \textit{i.e.},
\begin{equation*}
\text{Krum}(X)=x_{i^*} \quad \text{where} \quad i^* = \argmin_{i\in[n]} \sum_{j\in C_i} \lVert x_i-x_j\rVert^2.
\end{equation*}

\paragraph{Geometric median (GM) \cite{PKH22}} Given a set of $n$ vectors $X= \left\{x_i \in \mathbb{R}^d\right\}_{i=1}^n$, their geometric median, denoted by $\text{GM}(X)$, is defined to be a vector that minimizes the sum of Euclidean distances to all the vectors, \textit{i.e.},
\begin{equation*}
    \text{GM}(X) = \argmin_{x\in\mathbb{R}^d} \sum_{i=1}^n \lVert  x - x_i\rVert.
\end{equation*}

In general, there exists no closed form for GM.
However, an approximate solution can be derived by multiple iterations of the smoothed Weiszfeld's algorithm \cite{pillutla2022robust}, each of which has a time complexity of $\mathcal{O}(nd)$.

\paragraph{Centered clipping (CC) \cite{KHJ21}} Let $X= \left\{x_i \in \mathbb{R}^d\right\}_{i=1}^n$ be the set of vectors to be aggregated. Starting from some point $v_0$, one iteratively computes a sequence of vectors $v_m$, $m \in [M]$ according to
\begin{equation*}
    v_{m} = v_{m-1} + \frac{1}{n} \sum_{i=1}^n (x_i-v_{m-1}) \min \left\{ 1, \frac{\tau_m}{\lVert x_i-v_{m-1} \rVert}  \right\}
\end{equation*}
where $\tau_m\geq $ is a clipping parameter. Then, $\text{CC}(X)=v_M$.

CC is an iterative algorithm, and it can be considered scalable if the number of iterations $M$ is small enough \cite{KHJ21}. In particular, each CC update has a time complexity of $\mathcal{O}(nd)$. However, the value of $M$ that leads to a satisfactory aggregation depends on the input vectors $X$ and the clipping parameter.

The aggregation mechanisms listed are not exhaustive. There are other aggregation rules, such as the Minimum Enclosing Ball with Outliers \cite{yi2024near}, as well as variants that are modifications or combinations of the typical ones listed above.

{We remark that average aggregation, which is shown to be fragile to attacks, can be formulated as the following optimization problem
\begin{equation*}
    \text{Avg}(X) = \argmin_{x\in\mathbb{R}^d} \sum_{i=1}^n \lVert x - x_i \rVert^2.
\end{equation*}
We can see that Avg, CWMed, and GM solve a similar problem but use different norms to define the objective function. The robustness of CWMed and GM stems from the resilience of their norms to outliers, in contrast to the squared norm used by Avg, which is not robust to outliers.
}







To further improve the performance of resilient aggregation rules, pre-processing techniques can be employed. The idea is to transform the vectors to be aggregated into a new set, to which resilient aggregation is then applied.

\paragraph{Bucketing \cite{karimireddy2021byzantine}} 
The bucketing procedure finds its root in the median-of-means \cite{lugosi2019sub}. Given a set of $n$ vectors $X= \left\{x_i \in \mathbb{R}^d\right\}_{i=1}^n$ and an integer parameter $s \geq 1$, the $s$-bucketing procedure generates $\lceil \nicefrac{n}{s} \rceil$ vectors, where each vector is produced by the following steps:
\begin{itemize}
    \item[1)] Pick random permutation $\pi$ of $[n]$;
    \item[2)] Compute the following quantity
    \begin{equation*}
        y_i = \frac{1}{s}\sum_{j=(i-1)\cdot s +1}^{\min\{ n,i\cdot s \}} x_{\pi(j)}.
    \end{equation*}
\end{itemize}

\paragraph{Nearest neighbor mixing (NNM) \cite{AFGGPS23}} Given a set of $n$ vectors $X= \left\{x_i \in \mathbb{R}^d\right\}_{i=1}^n$, and the largest number of faulty nodes $n_f<\nicefrac{n}{2}$. NNM outputs $Y= \left\{y_i \in \mathbb{R}^d\right\}_{i=1}^n$,
where each $y_i$ is generated according to the steps:
\begin{itemize}
    \item[1)] Sort the vectors in $X$ to get $(x_{i:1},\dots,x_{i:n})$ such that
    \begin{equation*}
        \lVert x_{i:,1}-x_i \rVert \leq \dots \leq  \lVert x_{i:,n}-x_i \rVert;
    \end{equation*}
    \item[2)] Average the $n-n_f$  nearest neighbors of $x_i$, \textit{i.e.},
\begin{equation*}
    y_i = \frac{1}{n-n_f} \sum_{j=1}^{n-n_f} x_{i:j}.
\end{equation*}
\end{itemize}

Pre-aggregation techniques show improvements in both theory and practice when combined with most resilient aggregation mechanisms. Among these, NNM requires a higher computational load compared to bucketing.



\subsubsection{Resilient consensus over peer-to-peer topology}

Although resilient consensus in a peer-to-peer topology also requires each agent to aggregate messages from its neighbors, the resilient aggregation rules for federated topology reviewed in the previous section are not directly applicable for two reasons. First, in federated topology, resilient aggregation is typically performed by a coordinator that does not produce a model based on its own data. In contrast, agents in resilient consensus generate updates independently, which must be utilized. Second, the resilient aggregation rules do not meet the safety requirements in consensus as described in Section \ref{subsec:objective_shift}. In this section, we introduce resilient consensus protocols designed to address these two issues.

\paragraph{Mean subsequence reduced (MSR)} Given a set of $n$ \textit{scalars} $X= \left\{x_i \right\}_{i=1}^n \subset\mathbb{R}$, a reference point $x_j \in X$, and the maximum number of faulty nodes $n_f < \nicefrac{n}{2}$, MSR aims to remove some values from $X$ based on their relationship with the reference point $x_j$ and the value of $n_f$. The process is as follows:
\begin{itemize}
    \item[1)] Sort all values in $X$ in a descending order;
    \item[2)] If there are less than $n_f$ values in $X$ strictly larger than $x_j$, remove all values larger than $x_j$ from $X$. If there are at least $n_f$ values in $X$ strictly larger than $x_j$, remove the $n_f$ largest values from $X$ (breaking ties arbitrarily);
   \item[3)] If there are less than $n_f$ values in $X$ strictly smaller than $x_j$, remove all values smaller than $x_j$ from $X$. If there are at least $n_f$ values in $X$ strictly smaller than $x_j$, remove the $n_f$ smallest values from $X$ (breaking ties arbitrarily).
   \end{itemize}
Denote the set of removed values as $R_j$ when taking $x_j$ as the reference point. Then, MSR outputs the following scalar
\begin{equation*}
    \text{MSR}(X,x_j) =  \frac{1}{\lvert  X \setminus R_j \rvert}\sum_{i\in X \setminus R_j} x_i.
\end{equation*}

\paragraph{Safe point-based aggregation} 
The extension of MSR to the multi-dimensional case is highly nontrivial, as running MSR for each coordinate does not guarantee the safety condition in Section \ref{subsec:objective_shift}, as illustrated by the following example.

\begin{example}[Resilient consensus of probability vectors \cite{vaidya2013byzantine}]
Consider a group of $4$ agents, where the honest agents ${1,2,3}$ have initial states represented by the following probability vectors:
\begin{equation*}
    x_1^{(0)} =\begin{bmatrix}
    a \\  b \\  b
    \end{bmatrix},     x_2^{(0)} =\begin{bmatrix}
    b \\  a \\  b
    \end{bmatrix},  x_3^{(0)} =\begin{bmatrix}
   b \\  b \\  a
    \end{bmatrix},
\end{equation*} 
where $1\geq a>b\geq 0$ and $a+2b=1$.
The fourth agent is Byzantine.
By running the MSR algorithm independently for each coordinate of the state, the honest agents end up agreeing on $\hat{x}=[b;b;b]$. Although each coordinate of $\hat{x}$ satisfies the scalar safety condition along each dimension separately, $\hat{x}$ does not fulfill the safety condition in the multi-dimensional case. To see this, observe that $\hat{x}$ is not a probability vector because $3b<a+2b=1$. Consequently, $\hat{x}$ is not contained within the convex hull of $x_1^{(0)}$, $x_2^{(0)}$ and $x_3^{(0)}$.
\end{example}

Motivated by this, the safe point-based aggregation strategy was developed. To proceed, we recall the definition of $n_f$-safe point.
\begin{definition}[{$n_f$-safe point}]\label{def:nf-safe-point}
    Given a set of $n$ points in $\mathbb{R}^d$, of which at most $n_f$ are adversarial, then a point in $\mathbb{R}^d$ that is guaranteed to lie in the interior of the convex hull of $n-n_f$ normal points is called an $n_f$-safe point.
\end{definition}

The idea of safe point-based aggregation mechanisms is that, at time step $t$, once an $n_f$-safe point $s_i^{(t)}$ is secured, each agent $i$ updates its state by
\begin{equation*}
    x_i^{(t+1)} =  x_i^{(t)}+ \alpha_i^{(t)} ( s_i^{(t)}-x_i^{(t)}),
\end{equation*}
where $\alpha_i^{(t)}$ is a possibly time-varying weight chosen from $(0,1)$. Since both $x_i^{(t)}$ and $s_i^{(t)}$ are in the convex hull of states of honest agents, safety is guaranteed. Next, two strategies are introduced to obtain an $n_f$-safe point.

\begin{figure*}[t]
\centering
\begin{subfigure}{.33\textwidth}
\centering
    \includegraphics[scale=0.65]{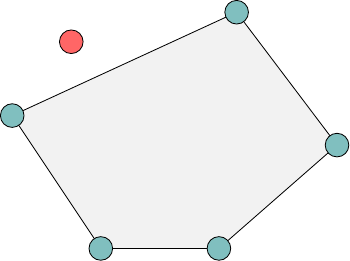}
    \caption{}
\end{subfigure}%
\begin{subfigure}{.33\textwidth}
\centering
    \includegraphics[scale=0.65]{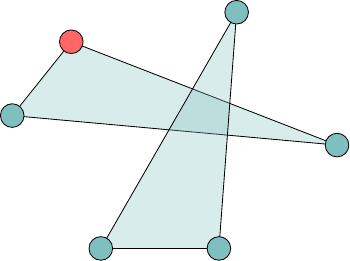}
    \caption{}
\end{subfigure}%
\begin{subfigure}{.33\textwidth}
\centering
    \includegraphics[scale=0.65]{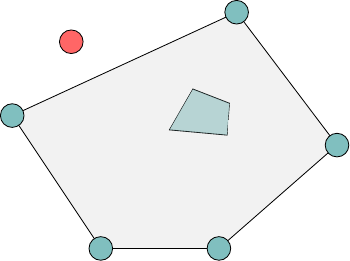}
    \caption{}
\end{subfigure}
\caption{A depiction of Tverberg partition for the case with $n=6$, $n_f=1$, and $d=2$. (a) plots the convex hull of the five points from honest agents; (b) presents a Tverberg partition consisting of two subsets, where only one of them contains adversarial agent and the intersection of the subsets is a Tverberg region; (c) illustrates that such Tverberg region is a subset of the convex hull of honest agents.}
\label{fig:Tverberg}
\end{figure*}

\begin{figure*}[t]
\centering
\begin{subfigure}{.33\textwidth}
\centering
    \includegraphics[scale=0.65]{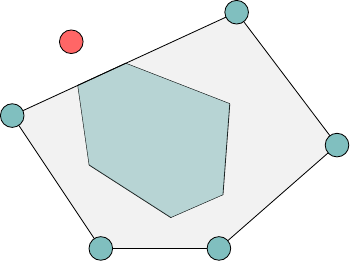}
    \caption{}
\end{subfigure}%
\begin{subfigure}{.33\textwidth}
\centering
    \includegraphics[scale=0.65]{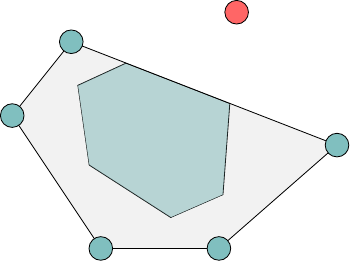}
    \caption{}
\end{subfigure}%
\begin{subfigure}{.33\textwidth}
\centering
    \includegraphics[scale=0.65]{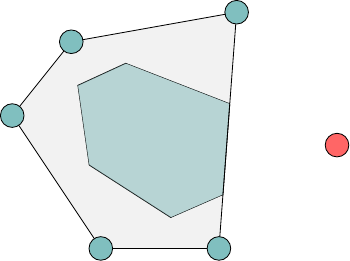}
    \caption{}
\end{subfigure}%
\qquad 
\begin{subfigure}{.33\textwidth}
\centering
    \includegraphics[scale=0.65]{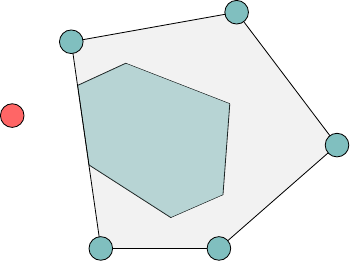}
    \caption{}
\end{subfigure}%
\begin{subfigure}{.33\textwidth}
\centering
    \includegraphics[scale=0.65]{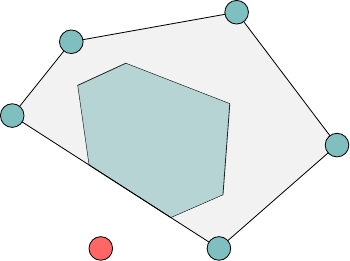}
    \caption{}
\end{subfigure}%
\begin{subfigure}{.33\textwidth}
\centering
    \includegraphics[scale=0.65]{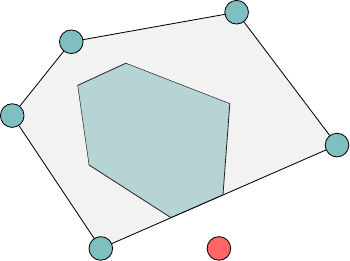}
    \caption{}
\end{subfigure}%
\caption{A depiction of the centerpoint region for the case where $n=6$, $n_f=1$, and $d=2$. The region of centerpoints is shaded in teal. Any point within this teal region is also a $1$-safe point, meaning that regardless of which one of the six agents is adversarial, every point in the teal region is guaranteed to lie within the convex hull formed by the remaining five honest agents.}
\label{fig:centerpoint}
\end{figure*}

\begin{itemize}
    \item \textit{Tverberg partition}: The idea is to partition a set of $n$ points in $\mathbb{R}^d$ into $n_f+1$ subsets such that the convex hulls of any two subsets intersect. The intersection of the convex hulls of all $n_f+1$ subsets is called a Tverberg region. Any point lying in the interior of this Tverberg region is a $n_f$-safe point by construction \cite{vaidya2013byzantine}. However, computing a Tverberg partition is NP-hard in general \cite{har2020improved}. Approximate algorithms can be employed to compute a partition \cite{mulzer2013approximating}, albeit at the cost of a stricter bound on the number of malicious agents \cite{park2017fault}. Figure~\ref{fig:Tverberg} provides an illustration of this technique.

    \item \textit{Centerpoint}:
    The notion of a centerpoint can be viewed as a generalization of the median in higher-dimensional Euclidean space \cite{abbas2022resilient}. Formally, given a set $X$ of $n$ points in $\mathbb{R}^d$, a centerpoint is a point (not necessarily in $X$) such that any closed half-space of $\mathbb{R}^d$ containing this point also contains at least $\nicefrac{n}{d+1}$ points from $X$. Figure~\ref{fig:centerpoint} provides an illustration of this technique. It was shown that an $n_f$-safe point is equivalent to an interior centerpoint if $n>n_f(d+1)$ \cite{abbas2022resilient}. For the special cases of $d=2$ and $d=3$, efficient computation methods for centerpoints were reported in \cite{mukhopadhyay1994computing} and \cite{chan2004optimal}, respectively. Their algorithms have time complexities of $\mathcal{O}(n)$ and $\mathcal{O}(n^2)$ for $d=2$ and $d=3$, respectively. For higher dimensions, \cite{miller2010approximate} provided a general approximate method. In all cases, these centerpoint-based approaches achieve a better trade-off between resilience and computational complexity than Tverberg partition-based ones.
\end{itemize}



\subsection{Metrics of resilience}
\label{sec:metrics_of_resilience}

Analyzing the resilience of decentralized algorithms from a theoretical standpoint requires the definition of suitable metrics, which we review in the following.

\subsubsection{Aggregation resilience}

To quantify aggregation resilience, we introduce the concept of concentration, which refers to the property that, for any subset of inputs of size $n-n_f$, the output of the aggregation rule is in close proximity to the average of these inputs \cite{FGGPS22,KHJ22,guerraoui2023byzantine}.

\begin{definition}[Concentration]
For $n_f<n$, a resilient aggregation rule $A$ is said to satisfy concentration if there exists a real-valued function $\Phi$ such that for any set of $n$ vectors $X := \left\{x_i\right\}_{i=1}^n$ and any subset $S\subseteq[n]$ with $\lvert S \rvert=n-n_f$,
    \begin{align}
        \expec{}{\norm{A(X) - \overline{x}_{S}}^2} \leq  \expec{}{\Phi(x_i,i\in S)},
    \end{align}
     where $\overline{x}_S := (n-n_f)^{-1}\sum_{i\in S} {x_i}$ and the expectation is taken over the randomness of the input vectors.
\end{definition}

The concentration criteria in the literature differ from each other in the definition of proximity, \textit{i.e.}, $\Phi$. It serves as quantitative conditions and can therefore be used for analytical purposes. Next, we introduce three common concentration conditions, that is, $(n_f,\lambda)$-resilient averaging~\cite{FGGPS22}, $(\delta_{\text{max}},\zeta)$-agnostic robust aggregator (ARAgg)~\cite{KHJ22}, and $(n_f,\kappa)$-robustness~\cite{AFGGPS23}.

\begin{itemize}
    \item \textit{$(n_f,\lambda)$-resilient averaging}: 
$
      \expec{}{\Phi(x_i,i\in S)} = \lambda \max_{i,j\in S}\norm{x_i - x_j}
$
  where $\lambda \geq 0$ is a parameter.
    \item \textit{$(\delta_{\text{max}},\zeta)$-ARAgg}: Suppose $\nicefrac{n_f}{n} \leq \delta_{\max} < 0.5$ and $\expec{}{\norm{x_i-x_j}^2} \leq \rho^2$ for all $i,j\in S$. Let $
       \expec{}{\Phi(x_i,i\in S)} =  \nicefrac{\zeta \rho^2 {n_f}}{n}$, where $\zeta \geq 0$ is a parameter.
       \item \textit{$(n_f,\kappa)$-robustness}: $
      \expec{}{\Phi(x_i,i\in S)} = {\kappa}{\sz{S}}^{-1}\sum_{i\in S}\norm{x_i - \overline{x}_S}^2
$
  where $\kappa \geq 0$ is a parameter.
\end{itemize}

\subsubsection{Graph robustness}

For algorithms applied over sparse network topologies, there is a close relationship between the topology and the maximum number of tolerable faulty agents. Graph robustness is a useful tool to reveal their coupling. It was first formalized in \cite{LeBlanc2013} and is recalled below:
\begin{definition}(Graph robustness):
	A {digraph} $\mathcal{G}=\{\mathcal{V},\mathcal{E}\}$ is \textit{$r$-robust}, if for any pair of disjoint and nonempty subsets $\mathcal{V}_1, \mathcal{V}_2\subsetneq\mathcal{V}$, at least one of the following statements hold:
	\begin{enumerate}
		\item There exists an agent in $\mathcal{V}_1$ such that it has at least $r$ in-neighbors outside $\mathcal{V}_1$;
		\item There exists an agent in $\mathcal{V}_2$ such that it has at least $r$ in-neighbors outside $\mathcal{V}_2$.
	\end{enumerate}

    {Additionally, the digraph $\calG$ is \textit{($r,s$)-robust} if any statement among 1., 2. and 3. (below) holds:
    \begin{enumerate}
    \setcounter{enumi}{2}
    \item There are no less than $s$ agents in $\mathcal{V}_1\cup\mathcal{V}_2$ such that each of them has at least $r$ in-neighbors outside their respective sets.
    \end{enumerate}
    }
\end{definition}


Intuitively speaking, graph robustness is a measure of connectivity in graphs. It asserts that for any two disjoint and non-empty subsets of nodes, there exist a significant number of nodes within these subsets that have a sufficient number of incoming connections from nodes outside the subsets. In other words, graph robustness ensures that each subset of nodes is adequately connected to the rest of the graph, reducing the likelihood of disconnected components or isolated clusters.
This requirement, along with resilient consensus protocols that eliminate a number of suspicious links, helps maintain connectivity and ultimately achieve agreement \cite{LeBlanc2013,Sundaram2019DistributedOU}.

\subsubsection{Cost redundancy} Because the identities of Byzantine agents are unknown, solving \eqref{eq:problem_DO_honest} is generally impossible. To circumvent this issue, researchers have explored additional conditions on the cost functions. A line of work has proposed the notion of redundancy in cost functions for the solvability of \eqref{eq:problem_DO_honest}, as introduced below.

\begin{definition}[$2n_f$-redundancy]
For a given set of honest agents $\mathcal{H}$, their costs are said to satisfy $2n_f$-redundancy 
    if for any subset $S \subset \mathcal{V}$ with $\lvert S \rvert \geq n-2n_f$,
    \begin{equation*}
        \argmin_{x}  \sum_{i\in S} f_i(x) =  \argmin_{x}  \sum_{i\in \mathcal{H}} f_i(x).
    \end{equation*}
\end{definition}
The $2n_f$-redundancy property implies that the sum of the costs of any $n - 2n_f$ honest agents and the sum of the costs of all honest agents share common minimizers. \cite{gupta2020resilience} demonstrated that this property is a necessary condition for exactly solving \eqref{eq:problem_DO_honest} in the presence of up to $n_f$ Byzantine agents. However, this condition is challenging to fulfill. Motivated by this difficulty, \cite{liu2021approximate} generalized it to the notion of $(2n_f,\epsilon)$-redundancy.

\begin{definition}[$(2n_f,\epsilon)$-redundancy]
For $\epsilon\geq 0$, the agents' local cost functions are said to satisfy the $(n_f,\epsilon)$-redundancy property if and only if for every pair of subsets of agents $S_1,S_2\subseteq [n]$, where $\lvert  S_1 \rvert=n-f$, $\lvert  S_2 \rvert\geq n-2f$ and $S_2\subseteq S_1$,
\begin{equation*}
    \text{dist}\left( \argmin_x \sum_{i\in S_1} f_i(x), \argmin_{x} \sum_{i\in S_2} f_i(x) \right) \leq \epsilon.
\end{equation*}
where $\text{dist}(X,Y)$ denotes the Hausdorff distance between two sets $X$ and $Y$, defined as
\begin{equation*}
     \text{dist}(X,Y):= \max \left\{  \sup_{x\in X} \inf_{y\in Y} \, \lVert x-y\rVert,  \, \sup_{y\in Y} \inf_{x\in X} \, \lVert y-x\rVert \right\}.
\end{equation*}
\end{definition}
\cite{liu2021approximate} demonstrated that $(2n_f,\epsilon)$-redundancy is both necessary and sufficient for a deterministic algorithm to output a point within a neighborhood around a true minimum point with radius $\epsilon$.

\smallskip

In the following sections we survey the Byzantine-resilient algorithms that have been proposed in recent years. These methods employ the resilient aggregation techniques discussed in section~\ref{subsec:resilient-aggregation}, and their performance is analyzed through the lens of the metrics defined in section~\ref{sec:metrics_of_resilience}.

\subsection{Resilient federated optimization}

\subsubsection{Homogeneous data}
Early Byzantine-resilient federated learning algorithms, such as those discussed in \cite{alistarh2018byzantine,AELA21}, which address convex and non-convex costs respectively, were built on stochastic gradient descent.
    In particular, a detection and isolation process guided by concentration inequalities was developed to reduce the sample and time complexity.
    The work in \cite{NIPS2017_f4b9ec30} focused on the development of the resilient aggregation rule, Krum, to alleviate the effect of faulty agents.
    \cite{el2021distributed} discovered that reducing the variance of honest gradients is beneficial to defending faulty agents, and proposed to apply aggregation rules to momentums to improve the performance.
    The paper~\cite{FGGPS22} coined the concept of $(f,\lambda)$-resilient averaging aggregators, and quantified the performance of several common aggregators under this framework.

\subsubsection{Heterogeneous data}
       Resilient federated learning with heterogeneous data distributions is arguably more challenging, since the distribution difference among agents makes the faulty agents harder to identify~\cite{KHJ21,KHJ22}. \cite{KHJ21} motivated the use of momentums in the heterogeneous case, and developed the concept of $(\delta_{\text{max}},\zeta)$-ARAgg and the CC rule for resilient aggregation. Further, \cite{KHJ22} presented a pre-aggregation technique called bucketing, which enhances the performance of aggregators within the framework of $(\delta_{\text{max}}, \zeta)$-ARAgg. Other criteria for evaluating resilient aggregators include the $(f, \kappa)$-robustness~\cite{AFGGPS23} and the $(f, \xi)$-robust averaging~\cite{AGGPS23}, with their relationships also discussed. \cite{wu2020federated} highlighted that variance reduction techniques lower the stochastic noise in honest gradients, which in turn improves the practical performance of resilient aggregators. For online learning problems, \cite{dong2023byzantine} showed that a linear regret in the fully adversarial case may be unavoidable, and proved a sublinear regret for an algorithm based on GM and the momentum method when honest agents have i.i.d. losses. \cite{YCKB18, ZWPWJ+23} discussed the optimal statistical rates achievable by resilient federated learning algorithms and presented nearly optimal algorithms for strongly convex loss functions. \cite{allouah2024robust} adopted another inequality to more accurately capture the heterogeneity in practice, under which a tighter convergence result can be stated.

Considering the case with both Byzantine attacks and communication compression, \cite{zhu2023byzantine_TSP} demonstrated that the Byzantine attacks and stochastic gradients make the compression noise more challenging to handle than attack-free cases. Motivated by this, the authors proposed to compress the gradient difference, which reduces both the compression and stochastic noises. 
This idea was further explored in \cite{gorbunov2022variance}, where improved convergence guarantees were established under relaxed assumptions. For similar problem setups, \cite{rammal2023communication} devised new algorithms with error feedback to accommodate biased compressors, \textit{e.g.}, $\text{Top}_k$ sparsification, to improve the practical performance. \cite{allouah2024tackling,malinovsky2023byzantine} investigated the behavior of federated learning with partial participation in the presence of Byzantine agents, and tailored aggregation mechanisms to tackle the challenge.

In another line of research, \cite{gupta2020resilience} coined the notion of redundancy in objective functions (later extended and relaxed in \cite{liu2021approximate}) and used it to develop the condition that ensures full resilience in decentralized optimization, that is, convergence to the consensual optimum of regular agents.

\subsection{Resilient distributed consensus and optimization}

We review resilient distributed algorithms, categorizing them into two primary objectives: consensus and optimization.

\begin{table*}[t]
\centering
\caption{A comparison of resilient consensus algorithms.}
\label{tab:comparison-resilient-consensus}
    \begin{tabular}{@{}ccccccc@{}}
    \hline
    \multirow{2}{*}{ [Ref.]}   & {Vector} & Attack$^\dagger$  & Aggregation & {Sparse} & \multicolumn{2}{c}{ Sufficient conditions}   \\ 
     \cline{6-7} 
 & consensus & model & rule & network   & Synchronous & Asynchronous   \\
    \hline
            \cite{LeBlanc2013}  & \xmark & B & \thead{weighted \\
MSR} &  \cmark & \thead{The network of honest agents \\is $(n_f+1)$-robust}  &  -   \\ 
            \cite{haseltalab2015approximate,haseltalab2015convergence}  & \xmark & B & \thead{weighted\\
MSR} &  \cmark &   \multicolumn{2}{c}{The network of honest agents is $(n_f+1)$-robust}    \\ 
            \cite{{ramos2023discrete}}  & \xmark & M & \thead{reputation-based \\
consensus} &  \cmark & \multicolumn{2}{c}{ $n_{f_i} < \frac{\lvert \mathcal{N}_i^- \rvert}{2}$ and $\lvert \mathcal{N}_i^- \rvert >2$}    \\ 
            \cite{vaidya2013byzantine,mendes2015multidimensional}  & \cmark & B & \thead{Tverberg partition} &  \xmark &  { $n> \max \{ 3n_f,(d+1)n_f \}$} & { $n> (d+2)n_f$}   \\ 
\cite{vaidya2014iterative,park2017fault}   & \cmark & B & \thead{
Tverberg partition} &  \cmark & { $n_{f_i}  \leq   \frac{\lvert \mathcal{N}_i^- \rvert}{d+1} -1$ }  & -   \\ 
\cite{yan2022resilient}  & \cmark & B & \thead{
auxiliary point} &  \cmark & {$n_{f_i} \leq \frac{\lvert \mathcal{N}_i^- \rvert}{ d+1} -1$}  & -   \\ 
\cite{abbas2022resilient}  & \cmark & B & \thead{
centerpoint} &  \cmark & $n_{f_i} \leq  \lfloor  \frac{\lvert \mathcal{N}_i^- \rvert}{ d+1}  \rfloor $   & -   \\ 
      \hline
    \end{tabular}\\
    {\footnotesize 
$^\dagger$ M and B represents the malicious and Byzantine attack model, respectively. \\
}
\end{table*}

\subsubsection{Resilient consensus} The resilient consensus algorithms are reviewed for both scalar and vector scenarios. Table \ref{tab:comparison-resilient-consensus} provides an overview of these algorithms.

\paragraph{Scalar consensus} The early works on resilient \textit{scalar} consensus date back to \cite{dolev1986reaching,kieckhafer1992fault}. However, these were restricted to complete networks. Extensions to general sparse networks were developed in \cite{LeBlanc2013,zhang2015notion,dibaji2017resilient}, where the analysis was based on graph robustness. In these frameworks, the key module for achieving resilience against false data injection (FDI) attacks is the MSR. Under reasonable conditions on the graph, the honest agents using the MSR algorithm reach asymptotic consensus \cite{LeBlanc2013}.

For asynchronous networks, \cite{haseltalab2015approximate} derived necessary and sufficient topological conditions for achieving approximate consensus. Notably, the conditions in \cite{haseltalab2015approximate} are not more restrictive than those required for synchronous networks. The rate of convergence was examined in \cite{haseltalab2015convergence}. \cite{ramos2023discrete} proposed a novel approach to resilient consensus-seeking by integrating a reputation system into the algorithm. In their framework, each agent dynamically evaluates the reputation of its neighbors based on the information received from them over time. This reputation scoring mechanism enables agents to gradually identify and mitigate the influence of malicious neighbors. Under a set of well-defined conditions, including the assumptions that malicious agents do not perturb the initial variables and that the number of malicious neighbors for each agent is less than half, the proposed algorithm exhibits linear convergence to consensus. Notably, this convergence occurs while simultaneously identifying the malicious agents within the network. Based on evidence theory, \cite{bonagura2023resilient} developed a methodology where each agent computes reputation values for its neighbors, which is combined with a weight correction scheme to mitigate the influence of malicious agents during the consensus-seeking process. Note that the aforementioned works considered the compromised agents to be static throughout the consensus-seeking process. In contrast, \cite{Wang2022ResilientRC} explored a mobile attack model in which a fixed number of adversaries can dynamically switch their targets and developed modified MSR algorithms to mitigate the effects caused by such switching behavior.

\paragraph{Vector consensus} Extending the approach to the \textit{vector} case (\textit{i.e.}, $d\geq 2$) is highly non-trivial, as running MSR independently for each coordinate might cause honest agents to converge to a point outside the convex hull of their initial states \cite{vaidya2013byzantine}. Considering complete communication networks, \cite{vaidya2013byzantine} developed a resilient algorithm based on Tverberg partition \cite{soberon2012generalisation}, for which necessary and sufficient conditions were provided in both synchronous and asynchronous systems. The scenario with incomplete networks was examined in \cite{vaidya2014iterative}, where a sufficient condition, and a (different) necessary condition were given. \cite{mendes2015multidimensional} introduced the concept of safe area, which is guaranteed to be within the convex hull of non-faulty agents. They developed a resilient vector consensus scheme for complete networks. \cite{park2017fault} extended the safe point framework to incomplete networks and proposed a resilient rendezvous algorithm for multirobot systems, where the safe point is computed using an approximate algorithm for calculating a Tverberg partition. Another efficient algorithm to compute the safe point was developed in \cite{yan2022resilient}, where the time complexity is $\mathcal{O}((pr)^3)$ with $p=dn_f+1$ and $r=\dbinom{(d+1)n_f +1}{n_f}$. \cite{abbas2022resilient} interpreted the safe point as an interior centerpoint, based on which they achieved an improved trade-off between resilience and computation for safe point-based approaches. \cite{lee2024geometric} reported an extension of this method to handle cases with imprecise input data.

\subsubsection{Distributed optimization} Similar to consensus, we classify resilient distributed optimization algorithms according to their problem dimensions. Table \ref{tab:comparison-resilient-algorithms} summarizes these algorithms.

\begin{table*}[t]
\centering
\caption{A comparison of resilient distributed optimization algorithms for peer-to-peer topologies.}
\label{tab:comparison-resilient-algorithms}
    \begin{tabular}{@{}cccccc@{}}
    \hline
    \multirow{2}{*}{ [Ref.]}  & Vector  & Attack$^\dagger$  & Aggregation & {Aggregated }  & \multirow{2}{*}{Assumption}   \\
  & optimization  & model &  rule &  quantity &   \\
    \hline
    \cite{Sundaram2019DistributedOU}& \xmark  & B and M & MSR & \thead{decision  variables}  & $(2n_f+1)$-robust $\mathcal{G}$  \\ 
    \cite{su2020byzantine} & \xmark  & B & MSR & \thead{decision variables \\
    and gradients}  & \thead{every reduced graph has \\ a nonempty source component}\\ 
    \cite{zhao2019resilient} & \xmark  & B & \thead{averaging based on \\ honest neighbors' info} & \thead{decision variables}  & \thead{A known, connected network\\ of honest agents} \\
    \cite{wang2023resilient} & \xmark   & mobile M & MSR & \thead{decision  variables}  & \thead{$n\geq 4n_f+4$ and \\ $\lvert \mathcal{N}_i^- \rvert \geq 2n_f+1+\frac{n}{2} $}  \\ 
    \cite{yang2019byrdie} & \cmark  & B & \thead{coordinate-wise \\ MSR} &  \thead{decision  variables}  & \thead{every reduced graph has \\ a nonempty source component} \\
     \cite{fang2022bridge}  & \cmark  & B & general rules & \thead{decision  variables}  & \thead{every reduced graph has \\ a nonempty source component} \\
\cite{kuwaranancharoen2020byzantine,kuwaranancharoen2024scalable}  & \cmark  & B & \thead{trimmed mean and\\ distance-based rules} & \thead{decision and \\ auxiliary variables} & $(2n_f+1)$-robust $\mathcal{G}$\\
\cite{wu2022byzantine} & \cmark  &  B & \thead{iterative outlier\\ scissor (IOS)} & \thead{decision variables} & \thead{the network resulted from \\IOS is strongly connected}  \\
\cite{yang2024byzantine} &  \cmark  & B & remove-then-clip & \thead{decision variables} & \thead{bounded weights for faulty agents} \\
\cite{yu2023secure} & \cmark  & \thead{gradient \\ attack} & clip & \thead{decision variables} & \thead{common individual minimizer}\\
\cite{yemini2022resilience} &  \cmark  & B & \thead{averaging among \\ trusted neighborhood} & \thead{decision variables}  & \thead{ observations
of trust and honest \\ agents form a connected graph  } \\
\cite{zhu2022resilient} & \cmark  & B & \thead{resilient convex \\
combination \cite{wang2018resilient} } & \thead{decision variables} & \thead{cost and \\ graph redundancy} \\
            \cite{wang2022byzantine} & \cmark  & B & CWTM &  \thead{decision variables }   &  \thead{every reduced graph has \\ a nonempty source component}  \\
            
            \cite{wang2024d3} & \cmark  & B & general rules &  \thead{dual variables} &  \thead{ honest agents \\form a connected graph } \\ 
      \hline
    \end{tabular}
        \\
            {\footnotesize 
$^\dagger$ M and B represents the malicious and Byzantine attack model, respectively.
}
\end{table*}

\paragraph{Scalar optimization} \cite{Sundaram2019DistributedOU} studied the distributed scalar optimization problem in the presence of malicious agents. A resilient distributed optimization algorithm was developed based on the distributed subgradient method and the MSR. The authors proved that the estimates by regular agents converge to the convex hull of their local minimizers, under assumptions on the attack model and communication graph. Considering the Byzantine attack model and scalar problems, \cite{su2020byzantine} proposed a resilient distributed optimization algorithm based on trimmed mean. This algorithm guarantees convergence to a region characterized by the minimizers of the convex combinations of local objective functions of the honest agents, under proper conditions.
With the help of a set of identifiable trusted nodes, \cite{zhao2019resilient} developed an improved distributed scalar optimization algorithm against Byzantine attackers. To address scenarios with mobile malicious attackers, \cite{wang2023resilient} devised a distributed optimization algorithm based on the resilient consensus strategy proposed in their previous work \cite{Wang2022ResilientRC}. For constrained optimization, \cite{kaheni2022resilient} developed a resilient distributed projected subgradient method based on MSR, which guarantees convergence to a bound characterized by minimizers of costs from a subset of regular agents and convergence to the exact minimum when a certain cost redundancy condition is satisfied.

\paragraph{Vector optimization} To address multi-dimensional problems, \cite{yang2019byrdie} employed the framework of coordinate descent to decompose the problem into multiple one-dimensional subproblems. They proposed a multi-loop algorithm, where the one outer loop iterates over the dimensions, and the inner loop performs multiple iterations for each coordinate within the current dimension. To reduce sample complexity, \cite{fang2022bridge} proposed a loop-less algorithm compatible with several resilient aggregation rules. They established convergence guarantees for this algorithm, particularly when using the CWTM rule for general non-convex problems.
\cite{kuwaranancharoen2020byzantine,kuwaranancharoen2024scalable} introduced an auxiliary variable to each agent. This auxiliary variable is updated by resilient consensus with coordinate-wise MSR, and is used as a reference in the update of decision variables to defend Byzantine attackers. For stochastic optimization problems under the Byzantine attack model, \cite{wu2022byzantine} proposed a general robust aggregation rule. In this rule, each agent iteratively discards the message from its neighbors that deviates the most from the mean in this iteration. 
To more effectively filter Byzantine agents in distributed learning with homogeneous data, \cite{guo2021byzantine} devised a two-stage aggregation strategy. The first stage employs a principled neighbor selection approach, where each agent identifies a subset of neighbors based on the similarity between their respective states and those of their peers. In the subsequent stage, this neighbor set is refined through a discerning evaluation of the loss over the selected states. 
Recently, \cite{yang2024byzantine} proposed another two-stage aggregation strategy for Byzantine-resilient distributed learning. In the first stage, each agent filters its neighboring set by discarding nodes whose states significantly deviate from its own state. The second stage involves clipping the variables in the remaining neighborhood set before performing the aggregation.
Assuming agents share a common minimizer, \cite{yu2023secure} proposed a resilient distributed optimization algorithm that incorporates momentum methods and gradient clipping techniques. 
Leveraging inter-agent trust, \cite{yemini2022resilience} proposed a resilient distributed optimization algorithm with two key features: 1) enabling each agent to identify its malicious neighbors, and 2) ensuring convergence to the exact optimum despite the presence of malicious agents. Based on the notions of cost redundancy \cite{gupta2020resilience} and graph robustness, \cite{zhu2022resilient} developed a resilent algorithm that converges to the exact optimum.


\paragraph{Resource allocation} 
For a class of resource allocation problems where the coupled constraint is a function of the average of local decision variables, \cite{turan2020resilient} developed a federated primal-dual optimization algorithm that employs a CWMed-based mean estimation rule to achieve resilience.
For a class of resource allocation problems with simple inequality constraints, \cite{wang2022byzantine} proposed a fully distributed algorithm based on the primal-dual algorithm with a regularized Lagrangian and CWTM. The algorithm linearly converges to a neighborhood of the optimum due to the presence of Byzantine agents and the regularization. For coupling equality constraints, \cite{wang2024d3} designed a fully distributed resilient resource allocation algorithm that properly aggregates dual variables to achieve resilience. The algorithm was shown to converge as long as the aggregation rule satisfies a concentration property.

{
\subsection{Discussions}

\paragraph{Achieving resilience under data heterogeneity} In the context of federated learning with heterogeneous datasets, convergence to the exact solution of \eqref{eq:problem_DO_honest} is not possible. This is because the inherent hetergeneity among honest agents makes the Byzantine updates more difficult to exclude. Indeed, \cite{KHJ22} presented a lower bound result, indicating that the optimization error is proportional to the variance of individual gradients. However, this heterogeneity model is restrictive and may not cover certain basic learning problems, such as least-squares regression. To address this issue,  \cite{allouah2024robust} developed another gradient dissimilarity notion to characterize hetergeneity and established lower bound accordingly. The notion of cost redundancy is another interesting research direction related to this issue, which serves a condition for the solvability of \eqref{eq:problem_DO_honest}, as detailed in Section \ref{sec:metrics_of_resilience}.

\paragraph{The curse of dimensionality} Solving multi-dimensional resilient decentralized optimization problems is challenging for two main reasons. First, in resilient consensus problems, running scalar resilient algorithms for each coordinate can lead to violations of safety requirements. Second, resilient aggregation and consensus algorithms that are efficient in multi-dimensional settings typically require heavy computational loads (\textit{e.g.} sorting vectors), as discussed in previous sections. Aggregation mechanisms that are computationally lighter tend to be less resilient, often exhibiting larger parameters in the concentration criterion for aggregation resilience. This highlights an efficiency-resilience trade-off in resilient decentralized optimization and learning.
}



%% file: paper/conclusion.tex
\section{Conclusion and Outlook}\label{sec:conclusion}

We have presented a comprehensive survey of existing privacy-preserving and resilient decentralized optimization and learning algorithms, highlighting their strengths and limitations.

\subsection{Conclusion}
This survey underscores the critical importance of a systematic design approach that prioritizes resilience and privacy preservation for secure decentralized optimization and learning systems. While a fruitful set of existing results has partially addressed this by integrating defense modules like differential privacy, encryption, and secret sharing protocols for privacy preservation, as well as resilient aggregation mechanisms for resilience into standard decentralized frameworks, incorporating additional defense mechanisms inevitably impacts the system's overall utility. Notably, the interdependencies between these mechanisms and the dynamic systems can be leveraged to optimize the security-utility trade-off. Furthermore, different attacks pose varying consequences, necessitating distinct defense measures. Consequently, the deployment of defense mechanisms should be tailored to the specific criteria and practical requirements of the systems.



{
\subsection{Topics not covered in this survey}


In Section \ref{sec:intro}, we discuss the CIA triad in information technology systems: {confidentiality}, {integrity}, and {availability}.
The first two, confidentiality and integrity, pertain to data privacy and model security in decentralized optimization and learning, which are the focus of this work. However, the third property, \textit{availability} (\textit{e.g.}, protection against denial-of-service attacks), is not covered in this study.

Adversarial machine learning, which involves incorporating adversarial samples into the training set to enhance robustness \cite{goodfellow2014explaining,kurakin2016adversarial}, is not covered in this survey. While this approach has shown excellent performance, it may not effectively address model security attacks that can behave arbitrarily, a key focus of this survey.

Finally, this survey does not address the design of attacks, such as optimization-based data privacy attacks \cite{zhu2020deep} and gradient variance-based model security attacks \cite{baruch2019little}. However, the threat model we consider is general enough to encompass these attacks as special cases, and the secure decentralized optimization and learning algorithms reviewed in this survey are applicable to them.

}

\subsection{Future research directions}

Several important and challenging future research directions have recently garnered significant attention. Some of the most notable areas are summarized below.

\paragraph{Achieving privacy and resilience simultaneously} The majority of existing secure decentralized algorithms consider either privacy preservation or resilience, but not both. However, decentralized optimization and learning algorithms may face threats targeting both data and the algorithm itself. This challenging yet important problem has recently drawn attention from various communities, such as consensus \cite{fiore2019resilient,ramos4733736reputation} and decentralized machine learning \cite{AGGPS23,allouah2023can}. In particular, \cite{fiore2019resilient} combined the Laplace mechanism and MSR to make the consensus algorithm privacy-preserving and resilient, while \cite{ramos4733736reputation} incorporated the Gaussian mechanism into a reputation-based resilient consensus algorithm \cite{ramos2023discrete} to achieve DP. For secure federated learning, \cite{AGGPS23} showed that there is a fundamental accuracy cost when simultaneously enforcing DP and resilience. This is primarily because DP mechanisms introduce artificial data heterogeneity among workers through noise injection, making identifying malicious agents more challenging. However, this research direction is far from complete. For example, combining other privacy-preserving mechanisms, such as HE and SS, with resilient aggregation to achieve privacy and resilience simultaneously is worthy of exploration. Investigating the fundamental performance limits of secure distributed optimization and learning is also an interesting avenue. Another intriguing direction is to combine privacy and resilience when considering weaker yet meaningful adversaries, such as the honest non-curious coordinator scenario \cite{kairouz2021distributed}.

\paragraph{Securing decentralized algorithms in asynchronous environment} Existing secure decentralized optimization algorithms implicitly assume synchronous communication. This assumption facilitates the implementation of resilient aggregation and provides a basis for theoretical justification of the algorithms. However, achieving global synchronization over a network can be challenging in practice. Furthermore, synchronized updates are inefficient and unreliable, as the time taken per iteration is determined by the slowest node and the optimization process is vulnerable to single-node failures \cite{wu2023delay,wu2023delay-ago}. Consequently, defending against attackers in an asynchronous environment has emerged as an important yet challenging task. 
A few recent attempts have been made in \cite{damaskinos2018asynchronous,yang2023buffered} for resilient federated learning, where their approaches to deal with asynchrony involve a dampening scheme that scales each gradient based on its staleness and creating a buffer at the coordinator while waiting for enough gradients to arrive. Interesting future research directions include the consideration of peer-to-peer topologies and more general models of asynchrony \cite{wu2023delay}.

\paragraph{Detecting and isolating faulty agents to improve accuracy}
A natural approach to improving the security-utility trade-off is to incorporate proactive fault detection and isolation mechanisms \cite{bianchin2015distributed} into decentralized optimization and learning systems. Recent works, such as \cite{ramos2023discrete}, implemented reputation scoring systems within consensus-seeking algorithms to address this. In federated learning, \cite{sun2023shapleyfl} adjusted the weights assigned to each agent based on their contributions, as measured by the Shapley value. Despite these advancements, it remains unclear how to effectively combine proactive mechanisms with passive robust aggregation techniques to achieve theoretical guarantees on system performance and resilience. Additionally, there is a need to explore whether these strategies can be extended to distributed optimization problems over peer-to-peer topologies.

\paragraph{Exploiting additional side information to enhance resilience} Most resilient decentralized optimization approaches rely on the transmitted data between agents to infer the presence of anomalies. This approach can be further strengthened by exploiting additional side information from the CPS, such as arriving directions of WiFi signals \cite{xiong2013securearray}, to provide supplementary avenues for resilience. For example, \cite{yemini2021characterizing,yemini2022resilience} developed an inter-agent trust evaluation framework that ensures convergence to the exact optimum in distributed optimization. Along this line of research, future efforts may include: 1) formalizing more characterizable notions of trust in multi-agent networks and 2) developing novel decentralized optimization algorithms that seamlessly integrate and exploit trust values.

\paragraph{Leveraging blockchain technology to secure decentralized learning} Blockchain technology, originally developed for the digital currency Bitcoin, combines cryptographic algorithms and decentralized consensus protocols to enable secure data sharing and storage among agents, even in the presence of faulty agents \cite{belotti2019vademecum}.
By design, it is a suitable architecture for secure decentralized optimization. Indeed, this architecture has paved the way for innovative applications in various domains. For example, \cite{miao2022privacy} developed a secure federated learning system, \cite{strobel2023robot} studied blockchain-based robot swarms, and \cite{chen2022blockchain} seamlessly combined blockchain and energy dispatch tasks through replacing the mathematical puzzle in Proof of Work by Proof of Solution to a task-related optimization problem. While the integration of blockchains into decentralized optimization and related problems holds significant promise, several technological challenges need to be addressed. For example, the computing, storage, and communication limitations of mobile multi-agent systems must be considered, as they may pose constraints on the effective implementation of blockchain-based solutions.

\paragraph{Decentralized optimization and learning for control} 
Decentralized optimization and learning are closely tied to the two primary tasks in control: system identification \cite{wang2023fedsysid, azzollini2023robust} and control design \cite{wang2023model, wangrobot}. Recent advancements in decentralized optimization have laid the foundations for advanced control design. For example, developing violation-free distributed optimization algorithms under coupling constraints for safe distributed control, as recently attempted in \cite{liu2024achieving,tan2024continuous}, is a key direction. Another promising future direction is secure distributed control based on decentralized optimization methodologies.
Finally, leveraging fast distributed optimization algorithms to enable the distributed implementation of data-enabled predictive control (DeePC) \cite{coulson2019data} is also worthy of exploration.